\documentclass[11pt,letterpaper]{article}
\usepackage[T1]{fontenc}
\usepackage{lmodern}
\usepackage[margin=1in]{geometry}
\usepackage[authoryear,round]{natbib}
\setcitestyle{citesep={;},aysep={,},yysep={;}}
\usepackage{amsmath,amssymb,amsthm,mathtools,bm,booktabs,array,microtype}
\usepackage{float}
\usepackage{xcolor}
\usepackage{url}
\usepackage{hyperref}
\hypersetup{
  colorlinks=true,
  allcolors=blue,
  pdftitle={Coordinatewise Epsilon for Adam under Heavy-Tailed Noise},
  pdfauthor={Jeonseong Kim and Chenglin Fan}
}

\newtheorem{theorem}{Theorem}[section]
\newtheorem{lemma}[theorem]{Lemma}
\newtheorem{proposition}[theorem]{Proposition}
\newtheorem{corollary}[theorem]{Corollary}

\theoremstyle{definition}

\newtheorem{assumption}[theorem]{Assumption}

\newcommand{\E}{\mathbb E}
\newcommand{\R}{\mathbb R}
\newcommand{\F}{\mathcal F}
\newcommand{\eps}{\varepsilon}
\newcommand{\grad}{\nabla}
\newcommand{\Diag}{\operatorname{Diag}}
\newcommand{\clip}{\operatorname{clip}}

\title{Coordinatewise Epsilon for Adam\\under Heavy-Tailed Noise}
\author{
Jeonseong Kim\\
Yonsei University\\
Seoul, Korea\\
\texttt{JeonseongKim@gmail.com}
\and
Chenglin Fan\\
Seoul National University\\
Seoul, Korea\\
\texttt{fanchenglin@snu.ac.kr}
}

\date{}

\begin{document}
\maketitle
\pagestyle{plain}
\thispagestyle{plain}

\begin{abstract}
Adam estimates second moments coordinatewise, but typically adds the same scalar $\eps$ to every denominator. We study how this shared $\eps$ affects convergence when noise and curvature vary across coordinates. Under coordinatewise finite-$p$ noise moments, $p\in(1,2]$, choosing $\eps_i$ from the noise and curvature bounds for each coordinate attains the heavy-tailed $\ell_1$-stationarity benchmark $O(T^{-\frac{p-1}{3p-2}})$, matching the known optimal time exponent for smooth nonconvex stochastic optimization. For $1<p<2$, Adam with a shared $\eps_t$ admits a lower bound of order $\Omega(T^{-\frac{p-1}{2p}})$. This lower bound holds for every predictable scalar choice of $\eta_t$, $\beta_{1,t}$, $\beta_{2,t}$, and $\eps_t$, precluding a dimension-uniform benchmark rate for this class.
\end{abstract}

\section{Introduction}

Adam and AdamW~\citep{kingma2015adam,loshchilov2019decoupled} are standard optimizers in large-scale language-model training~\citep{brown2020language,touvron2023llama}. Heavy-tailed stochastic-gradient noise has been observed in deep networks and attention-model training~\citep{simsekli2019tail,zhang2020attention}, including recent language-model pretraining runs~\citep{yu2026sign}. For linear models with cross-entropy loss, class imbalance produces heterogeneous gradients and Hessians~\citep{kunstner2024class}. These observations motivate studying how Adam's coordinatewise normalization behaves when noise and curvature vary across coordinates.

Adam estimates the second moment separately in each coordinate. Its update adds the same scalar to every denominator:
\[
x_{t+1,i}=x_{t,i}-\eta_t
\frac{m_{t,i}}{\sqrt{v_{t,i}}+\eps_t}.
\]
The additive term $\eps_t$ prevents small denominators and determines how strongly the update responds to the second-moment estimate~\citep{zaheer2018adaptive,tong2022calibrating}. If $\eps_t$ dominates, differences between the second-moment estimates have little effect on the denominators. If $\eps_t$ is small, those differences directly shape the updates.

\citet[Eq.~(2)]{yu2026openproblem} ask whether AdamW can attain the heavy-tailed $\ell_1$-stationarity benchmark of sign-based methods. We adopt the same curvature and noise assumptions as their gradient-independent setting, with one stochastic gradient per step and zero weight decay. Write $L=\sum_iL_i$ for the total coordinatewise curvature bound and $\Sigma=\sum_i\sigma_i$ for the total finite-$p$ noise scale, where $p\in(1,2]$. We compare shared and coordinatewise choices of $\eps$ at fixed $L,\Sigma$, allowing the dimension and the allocation of these scales to vary. Our target is their benchmark uniformly over these choices. We refer to $T^{-\frac{p-1}{3p-2}}$ as the sharp heavy-tailed rate, since this time exponent is known to be optimal for smooth nonconvex stochastic optimization~\citep{liu2025heavy}.

For $1<p<2$, Adam with a shared $\eps_t$ has a lower bound with the slower exponent $(p-1)/(2p)$ on instances whose dimension grows with $T$, even when all scalar hyperparameters may depend on the past. In contrast, choosing $\eps_i$ according to the coordinatewise noise and curvature bounds attains the benchmark for every $p\in(1,2]$.

The upper bound balances two goals: choosing $\eps_i$ large enough to control momentum noise and gradient drift in each coordinate, while keeping $\sum_i\eps_i$ small. Given the coordinatewise bounds, we take, with $\rho_2=1-\beta_2$,
\begin{equation}
\eps_i
\asymp
\rho_2^{-\frac{2-p}{2p}}
\max\left\{\sigma_i,\frac{L_i}{L}\Sigma\right\}.
\label{eq:coordinate-floor-intro}
\end{equation}
The maximum combines the noise and curvature requirements, and
$\sum_i\max\left\{\sigma_i,\frac{L_i}{L}\Sigma\right\}\le2\Sigma$ controls the total.

The lower-bound construction pairs coordinates and adds a control coordinate. On an event of constant probability, the control coordinate offsets the progress of all pairs, keeping the iterates in a region with constant nonzero gradient.

\paragraph{Contributions.}
\begin{enumerate}
\item \textbf{Coordinatewise upper bound.} Theorem~\ref{thm:vector-upper} attains the benchmark dependence on $\Delta,L,\Sigma,T$, and in particular matches the known optimal heavy-tailed time exponent. A self-normalized tracking estimate controls heavy-tailed momentum noise while retaining the random denominator for large observations.
\item \textbf{Lower bound for a shared $\eps$.} Theorem~\ref{thm:scalar-impossibility} gives a lower bound for $1<p\le2$ through a pathwise comparison of normalized updates. The oracle outputs are gradients of scalar sample functions.

\end{enumerate}

\section{Related work}

\paragraph{Adaptive normalization and Adam theory.}
AdaGrad introduced diagonal scaling by accumulated coordinatewise gradients; Adam uses exponential moving averages to estimate first and second moments~\citep{duchi2011adaptive,kingma2015adam}. The classical counterexample of \citet{reddi2018convergence} concerns exponential forgetting and nonmonotone effective stepsizes. \citet{heilman2026adam} extend this phenomenon to fixed $\beta_1,\beta_2,\eps$, while \citet{zhou2018adashift} study same-time correlation between the gradient and its second-moment estimate. Convergence analyses address bounded gradients, finite or affine variance, generalized smoothness, and suitable schedules~\citep{chen2019convergence,defossez2022simple,zhang2022adam,hong2024convergence}. Under Euclidean smoothness and coordinatewise bounded variance, \citet{li2025adamw} establish an $O(\sqrt d\,T^{-1/4})$ expected $\ell_1$-gradient rate for AdamW, with Adam as a special case.

\citet{liu2025anisotropic} analyze AdaGrad under anisotropic smoothness and noise conditions, while \citet{jiang2025adagrad} combine coordinatewise smoothness and variance with $\ell_1$ stationarity to compare AdaGrad and SGD. Together, these works motivate expressing guarantees in a coordinatewise geometry and tracking how dimension enters when the coordinate scales are aggregated. We use this viewpoint to compare shared and coordinatewise choices of $\eps$ at fixed $L=\sum_iL_i$ and $\Sigma=\sum_i\sigma_i$.

\paragraph{Heavy-tailed nonconvex optimization.}
For $1<p<2$, a finite $p$th noise moment need not imply finite variance. Gradient clipping gives expectation and high-probability guarantees in this regime~\citep{zhang2020attention,cutkosky2021highprob}. \citet{zhang2020attention} establish the classical heavy-tailed time exponent $(p-1)/(3p-2)$ under a global-norm nonconvex model and also study coordinatewise clipping. Later work develops refined clipping and normalization methods~\citep{liu2023breaking,hubler2025normalization,liu2025heavy}. Models in this literature vary in whether moments are imposed on stochastic gradients or centered noise, and in whether control is global or coordinatewise. Sign-based methods provide the coordinatewise $\ell_1$ comparison relevant here~\citep{yu2026sign,yu2026openproblem}.

\paragraph{Adaptive methods under heavy-tailed noise.}
Heavy-tailed guarantees for adaptive methods cover several algorithms and stationarity notions. \citet{chezhegov2025clipping} show that explicit clipping improves the high-probability confidence dependence of Adam-Norm and AdaGrad-Norm. \citet{liu2026adagrad} proves an expected rate for coordinatewise AdaGrad when $p>4/3$ and gives an algorithm-dependent lower bound. Recent online-to-nonconvex analyses establish guarantees for Adam under relaxed stationarity criteria~\citep{liu2026adam,pang2026adam}. The setting considered here uses the expected $\ell_1$ norm of the true gradient, coordinatewise smoothness, and centered coordinatewise noise moments.

\paragraph{The choice of \texorpdfstring{$\eps$}{epsilon}.}
The effect of $\eps$ has also been studied: it can suppress adaptivity and enter convergence constants~\citep{zaheer2018adaptive}, while softplus transformations and alternative placements change anisotropic effective steps~\citep{tong2022calibrating,yuan2020eadam}. \citet{yu2026openproblem} formulate the AdamW benchmark and give a corridor obstruction under specified conditions on $\beta_1,\beta_2$ and the stepsize. This motivates treating $\eps$ as part of the optimizer's geometry and comparing a shared stabilization scale with coordinatewise ones.

\section{Preliminaries and problem setup}
\label{sec:setup}

\subsection{Notation and objective geometry}

For a positive definite matrix $D$, write $\|x\|_D=(x^\top D x)^{\frac{1}{2}}$; $\odot$ denotes coordinatewise multiplication. We use $a\lesssim b$ and $a\asymp b$ for one-sided and two-sided bounds up to positive constants. Unless stated otherwise, constants may depend on $p$ and the fixed compact set $\mathcal K$ of ratios $\rho_2/\rho_1$ introduced below, but not on $d,T,\Delta,L,\Sigma$ or the allocation of the coordinatewise scales. 

Let $f:\R^d\to\R$ be differentiable and lower bounded. Fix a diagonal geometry
\[
D_L=\Diag(L_1,\ldots,L_d),
\qquad L_i>0,
\qquad L=\sum_{i=1}^dL_i.
\]
\begin{assumption}[Coordinatewise smoothness]
\label{ass:gc}
We assume
\begin{equation}
\|\grad f(y)-\grad f(x)\|_{D_L^{-1}}
\le \|y-x\|_{D_L},
\qquad x,y\in\R^d.
\label{eq:gc}
\end{equation}
\end{assumption}
For twice continuously differentiable $f$, this condition is equivalent to
\[
-D_L\preceq\nabla^2 f(x)\preceq D_L,\qquad x\in\R^d.
\]
Assumption~\ref{ass:gc} implies the descent inequality
\begin{equation}
f(y)\le f(x)+\langle\grad f(x),y-x\rangle
+\frac12\|y-x\|_{D_L}^2.
\label{eq:smooth-descent}
\end{equation}
\begin{assumption}[Initial gap]
\label{ass:gap}
The initial point satisfies
\begin{equation}
f(x_1)-\inf_x f(x)\le\Delta.
\label{eq:gap}
\end{equation}
\end{assumption}

\subsection{Stochastic oracle and stationarity}

Let $\F_{t-1}$ denote the history before the $t$th oracle query, including the initialization, past oracle observations, and algorithmic randomness available before that query. We require $x_t$ and all time-$t$ parameters to be $\F_{t-1}$-measurable and call such parameter choices predictable.

\begin{assumption}[Finite-moment oracle]
\label{ass:oracle}
At time $t$, the oracle returns
\[
g_t=\grad f(x_t)+\xi_t,
\]
where, coordinatewise,
\begin{equation}
\E[\xi_{t,i}\mid\F_{t-1}]=0,
\qquad
\E[|\xi_{t,i}|^p\mid\F_{t-1}]\le\sigma_i^p,
\qquad p\in(1,2].
\label{eq:oracle}
\end{equation}
\end{assumption}
Write $\Sigma=\sum_i\sigma_i$ and assume $\Delta,L,\Sigma>0$.
The moment condition controls the centered noise and allows unbounded gradients. Coordinates may be dependent, and the noise distribution may depend on the past. At $p=2$ this is a coordinatewise conditional variance bound; for $p<2$ it permits infinite variance.

\paragraph{Stationarity criterion.}
We use the $\ell_1$-stationarity criterion in the open problem of \citet{yu2026openproblem},
\begin{equation}
R_T=\frac1T\sum_{t=1}^T\E\|\grad f(x_t)\|_1.
\label{eq:stationarity}
\end{equation}

\subsection{Adam families}

Starting from $m_0=v_0=0$, consider the family
\begin{align}
m_t&=\beta_{1,t}m_{t-1}+(1-\beta_{1,t})g_t,\\
v_t&=\beta_{2,t}v_{t-1}+(1-\beta_{2,t})g_t^{\odot2},\\
x_{t+1,i}&=x_{t,i}-\eta_t
\frac{m_{t,i}}{\sqrt{v_{t,i}}+\eps_{t,i}}.
\label{eq:adam-family}
\end{align}
The upper bound uses fixed $\eta_t\equiv\eta$,
$\beta_{k,t}\equiv\beta_k$, and $\eps_{t,i}\equiv\eps_i$;
write $\rho_k=1-\beta_k$.
For the lower bound, $\eta_t\ge0$, $\beta_{k,t}\in[0,1)$, and
$\eps_t\ge0$ may be arbitrary $\F_{t-1}$-measurable scalars, with the
restriction $\eps_{t,i}\equiv\eps_t$. This class includes
fixed-parameter Adam and standard scalar bias correction after its
deterministic factors are absorbed into $\eta_t$ and $\eps_t$.
We require updates to be finite and well defined almost surely on the problem class. We study the zero-weight-decay setting.

\section{Main results}
\label{sec:results}

\subsection{Upper bound with coordinatewise \texorpdfstring{$\eps_i$}{epsilon}}

Define
\begin{equation}
\rho_\star=
\left(\frac{\Delta L}{\Sigma^2T}\right)^{\frac{p}{3p-2}}.
\label{eq:target-scale}
\end{equation}

\paragraph{Choosing \texorpdfstring{$\eps_i$}{epsilon}.}
Write
\[
S=\sum_i\eps_i,\qquad
\kappa=\max_i\frac{L_i}{\eps_i}.
\]
For the stepsize
$\eta=\lambda\rho_1\Sigma\rho_2^{-\frac{2-p}{2p}}/L$,
with $\lambda>0$ sufficiently small, the upper proof uses three
sufficient requirements:
\begin{enumerate}
\item \emph{Momentum noise.} Each coordinate satisfies
\[
\eps_i\ge\rho_2^{-\frac{2-p}{2p}}\sigma_i.
\]
\item \emph{Gradient drift.} The updates satisfy
\[
\eta\kappa\le\lambda\rho_1.
\]
The coordinatewise bound
$\eps_i\ge\rho_2^{-\frac{2-p}{2p}}(L_i/L)\Sigma$
ensures this inequality and allows the drift contribution to be absorbed.
\item \emph{Total size.} The sum satisfies
\[
S\lesssim\rho_2^{-\frac{2-p}{2p}}\Sigma.
\]
After balancing the stepsize and averaging parameters, the proof gives
$R_T\lesssim\sqrt{\rho_1}\,S$.
\end{enumerate}
Taking the larger of the two coordinatewise lower bounds meets the first
two requirements. Their sums are both
$\rho_2^{-\frac{2-p}{2p}}\Sigma$, so this choice also meets the third.

\begin{theorem}[Coordinatewise choices of $\eps_i$ attain the sharp rate]
\label{thm:vector-upper}
Fix $p\in(1,2]$. Let $\mathcal K$ be a compact subset of
$(0,4/(2-p))$ when $p<2$, or of $(0,\infty)$ when $p=2$.
There exist constants $c,C,\lambda>0$, depending only on $p$ and
$\mathcal K$, such that the following holds whenever
$\rho_\star\le c$.
Choose
\[
\rho_1=\rho_\star,
\qquad
\frac{\rho_2}{\rho_1}\in\mathcal K,
\qquad
\eta=\lambda\frac{\rho_1}{L}\rho_2^{-\frac{2-p}{2p}}\Sigma,
\]
using the constant $\lambda$ above, and set
\begin{equation}
\eps_i=\rho_2^{-\frac{2-p}{2p}}
\max\left\{\sigma_i,\frac{L_i}{L}\Sigma\right\}.
\label{eq:coordinate-scales}
\end{equation}
Then
\begin{equation}
\frac1T\sum_{t=1}^T\E\|\grad f(x_t)\|_1
\le C
(\Delta L)^{\frac{p-1}{3p-2}}
\Sigma^{\frac{p}{3p-2}}T^{-\frac{p-1}{3p-2}}.
\label{eq:vector-rate}
\end{equation}
The bound is uniform over all objectives and oracles satisfying Assumptions~\ref{ass:gc}--\ref{ass:oracle}.
\end{theorem}

The choice \eqref{eq:coordinate-scales} gives
\begin{equation}
\sum_i\eps_i\le2\rho_2^{-\frac{2-p}{2p}}\Sigma.
\label{eq:floor-mass}
\end{equation}

\subsection{Lower bound for a shared \texorpdfstring{$\eps$}{epsilon}}

\begin{theorem}[Lower bound for a shared $\eps$]
\label{thm:scalar-impossibility}
Fix $1<p\le2$. There are constants $c,\delta_0>0$, depending only on $p$, such that for every horizon $T$, every rule in the predictable Adam class with a shared $\eps_t$, and every
\[
0<\delta\le\min\{\delta_0,T^{-\frac{p-1}{p}}\},
\]
there exist a dimension $d$, a convex continuously differentiable objective, an initial point, and a stochastic first-order oracle whose realizations are gradients of convex continuously differentiable scalar sample functions, satisfying
\[
f(x_1)-\inf f\le1,
\qquad
L\le1,
\qquad
\Sigma\le1,
\]
for which
\begin{equation}
\frac1T\sum_{t=1}^T\E\|\grad f(x_t)\|_1
\ge c\sqrt\delta.
\label{eq:shared-lb}
\end{equation}
For sufficiently large $T$, taking $\delta=T^{-\frac{p-1}{p}}$ gives a fixed-gap lower bound with exponent $(p-1)/(2p)$.
\end{theorem}

\subsection{Consequences}

\begin{corollary}[Shared $\eps$ under balanced coordinate scales]
\label{thm:balanced}
Under the assumptions of Theorem~\ref{thm:vector-upper}, suppose
\begin{equation}
\max_i\sigma_i\le C_0\frac{\Sigma}{d},
\qquad
\max_iL_i\le C_0\frac{L}{d}
\label{eq:balance}
\end{equation}
for a constant $C_0$. Set
\[
\eps_i\equiv\eps
=C_0\rho_2^{-\frac{2-p}{2p}}\frac{\Sigma}{d}.
\]
Then Adam with a shared $\eps$ satisfies, for a constant $C$ depending only on
$p$, $\mathcal K$, and $C_0$,
\[
R_T\le C
(\Delta L)^{\frac{p-1}{3p-2}}
\Sigma^{\frac{p}{3p-2}}T^{-\frac{p-1}{3p-2}}.
\]
\end{corollary}

The upper bound for arbitrary coordinate scales at fixed dimension is derived in Appendix~\ref{app:corollaries}.

\section{Proof overview}
\label{sec:proof-overview}

The upper bound uses the random denominator to control heavy-tailed momentum noise. The lower bound uses paired coordinates and a control coordinate to bound progress under shared scalar tuning. We give the key estimates here; the appendices contain the full proofs.

\subsection{Upper bound: controlling momentum noise}

Under the tuning of Theorem~\ref{thm:vector-upper}, the proof first bounds
the weighted gradient term, then the average denominator, and finally
the true gradient norm. Write $h_t=\grad f(x_t)$ and
$D_{t,i}=\sqrt{v_{t,i}}+\eps_i$.
With $S=\sum_i\eps_i$, the three key relations are
\begin{align}
\frac1T\sum_{t,i}\E\frac{h_{t,i}^2}{D_{t,i}}
&\lesssim\frac{\Delta}{\eta T}+\rho_1S,
\label{eq:overview-descent}\\
\frac1T\sum_{t,i}\E D_{t,i}
&\le2\left(\frac1T\sum_{t,i}\E\frac{h_{t,i}^2}{D_{t,i}}\right)+CS,
\label{eq:overview-denominator}\\
R_T^2&\le
\left(\frac1T\sum_{t,i}\E\frac{h_{t,i}^2}{D_{t,i}}\right)
\left(\frac1T\sum_{t,i}\E D_{t,i}\right)\nonumber\\
&\le
\left(\frac1T\sum_{t,i}\E\frac{h_{t,i}^2}{D_{t,i}}\right)
\left(2\frac1T\sum_{t,i}\E\frac{h_{t,i}^2}{D_{t,i}}+CS\right).
\label{eq:overview-stationarity}
\end{align}
The first follows from momentum tracking and smooth descent; the second
controls the denominators without an additional dimension factor; the
third is Cauchy--Schwarz over time, coordinates, and probability.
The prescribed parameters balance $\Delta/(\eta T)$ with $\rho_1S$.
Thus \eqref{eq:overview-descent}--\eqref{eq:overview-stationarity} give
\[
R_T^2\lesssim\rho_1S(\rho_1S+S)\lesssim\rho_1S^2,
\qquad R_T\lesssim\sqrt{\rho_1}\,S
\asymp\Sigma\rho_1^{\frac{p-1}{p}}.
\]

\paragraph{Tracking under heavy-tailed noise.}
The main difficulty in \eqref{eq:overview-descent} is controlling
$e_t=m_t-h_t$ when the noise may have infinite variance.
Fix a coordinate and suppress its index. The noise part of the first
moment is
\[
n_t=\rho_1\sum_{s=1}^t\beta_1^{t-s}\xi_s.
\]
The coordinatewise choice of $\eps$ permits the self-normalized estimate
\begin{equation}
\sum_{t=1}^T\E\frac{n_t^2}{D_t}
\le C\rho_1T\eps+
C\rho_1\rho_2\sum_{t=1}^T\E\frac{h_t^2}{D_t}.
\label{eq:main-noise-propagation}
\end{equation}
Its right-hand side contains the same weighted gradient term as
\eqref{eq:overview-descent}, with a coefficient that can be absorbed.
Initialization and gradient drift form the other parts of $e_t$.
Lemma~\ref{lem:self-normalization} controls them, using
$\eta\kappa\le\lambda\rho_1$ for the drift.

For $1<p<2$, the proof splits the noise at $\eps/\sqrt{\rho_2}$.
At this threshold, the term $\sqrt{\rho_2}|\xi_t|$ becomes comparable to $\eps$.
Below this threshold, truncated second moments and conditional centering
control the contribution. Above it, the random denominator controls
large observations: on $|\xi_t|>2|h_t|$,
\[
D_t\ge\frac{\sqrt{\rho_2}|\xi_t|}{2},\qquad
\frac{\xi_t^2}{\eps^{p-1}D_t^{2-p}}
\le\frac{C|\xi_t|^p}{\eps^{p-1}\rho_2^{(2-p)/2}}.
\]
On the complementary event, the bound contributes to the weighted
gradient term in \eqref{eq:main-noise-propagation}.

\paragraph{Propagation through the moment recursions.}
For $t\ge s$, $D_t\ge\beta_2^{(t-s)/2}D_s$.
Combined with the first-moment weights, this produces a geometric series
with ratio $\beta_1^2/\beta_2^{(2-p)/2}$. Write
\begin{equation}
\delta_p=1-\frac{\beta_1^2}{\beta_2^{(2-p)/2}}.
\label{eq:delta}
\end{equation}
For $\rho_2/\rho_1\in\mathcal K$ and sufficiently small $\rho_1$,
$\delta_p\asymp\rho_1>0$; the time sum is therefore at most $1/\delta_p$.
Appendix~\ref{app:selfnorm} gives the centered noise decomposition,
cross-term bounds, and full time summation proving
\eqref{eq:main-noise-propagation}. At $p=2$, the same estimate follows
directly from conditional variance and orthogonality.

\paragraph{The average denominator.}
To obtain \eqref{eq:overview-denominator}, the recursion bounds past
gradient contributions by the weighted gradient term in
\eqref{eq:overview-descent}.
The finite-$p$ noise moments contribute at most
$C\sigma_i\rho_2^{-(2-p)/(2p)}\le C\eps_i$ in coordinate $i$.
Summing over coordinates gives $CS$, controlled by
\eqref{eq:floor-mass}. Appendix~\ref{app:upper-proof} proves the descent and
denominator bounds in \eqref{eq:overview-descent}--\eqref{eq:overview-denominator}.

\subsection{Lower bound: a pathwise comparison}

We construct an objective
\[
f(z)=\phi(Y(z)),\qquad \phi(0)=0,\qquad \phi'(y)=\clip(y,0,1),
\]
where $Y$ is affine with a nonzero coefficient vector. On $Y(z)\ge1$,
the gradient of $f$ equals this fixed vector. Starting from $Y(z_1)=1$,
the construction maintains, on an event of constant probability,
\[
Y(z_{t+1})\ge Y(z_t)\ge1\qquad(1\le t\le T).
\]
This gives a lower bound on the gradient norm at every iterate.

The construction has three parts. A control coordinate receives
wrong-sign observations throughout this event and moves in a direction
that increases $Y$. Coordinate pairs have a uniformly bounded total
decrease under arbitrary shared averaging weights. Pairing makes this
bound depend only on the totals of the common first- and second-moment
weights, allowing the same comparison to cover arbitrary predictable
scalar tuning. The affine function $Y$ combines these contributions so
that the increase offsets the decrease.

\paragraph{Objective and coordinate scales.}
Fix $\delta$ as in Theorem~\ref{thm:scalar-impossibility}. With a fixed
positive constant $\gamma$ chosen in Appendix~\ref{app:lower}, set
\[
N=\lceil\delta^{-1}\rceil,
\qquad A=\frac{\gamma}{N},\qquad b=\frac{\gamma\sqrt\delta}{N}.
\]
Use one control coordinate $u$ and $N$ pairs $(w_{i,+},w_{i,-})$. With $s_i=(w_{i,+}+w_{i,-})/2$, define
\[
Y(z)=1+\gamma\delta\,u+b\sum_{i=1}^Ns_i.
\]
At $z_1=0$, $Y(z_1)=1$ and the initial gap is $1/2$.
On $Y(z)\ge1$, the gradient is the coefficient vector $\varrho$ of $Y$ and
\[
\|\grad f(z)\|_1=\gamma\delta+Nb
=\gamma(\delta+\sqrt\delta)\ge\gamma\sqrt\delta.
\]
\paragraph{Oracle and averaging weights.}
Let $\theta=1/(8T)$ and draw fresh randomness independently at each step. The control observation is $-\gamma\delta$ with probability $1-\theta$ and $\gamma\delta(2/\theta-1)$ otherwise. Each pair receives a uniformly random permutation of $\{A+b,-A\}$, using one common sign independent of the control draw. These observations form a vector $q_t$ with $\E q_t=\varrho$. Globally, set
\[
g_t(z)=\grad f(z)+q_t-\varrho
=\grad\bigl[f(z)+\langle q_t-\varrho,z\rangle\bigr].
\]
Let $\mathcal A$ be the event that all control observations take their common value; independent draws give $\Pr(\mathcal A)=(1-\theta)^T\ge7/8$.

Along any realized history, the recursions express $m_t$ and $v_t$ as nonnegative weighted sums of past gradients and squared gradients. Write the common scalar weights as $\alpha_{t,s},\nu_{t,s}$ and their totals as
\[
P_t=\sum_{s\le t}\alpha_{t,s},\qquad
V_t=\sum_{s\le t}\nu_{t,s}>0.
\]

\begin{lemma}[Sum of the two normalized updates]
\label{lem:paired-updates}
Suppose a coordinate pair receives a permutation of $\{A+b,-A\}$ at
each step up to time $t$. For $A,b>0$, the averaging weights above, and a shared
$\eps_t\ge0$, let $D_{t,i}^{\pm}=\sqrt{v_{t,i}^{\pm}}+\eps_t$. Then
\begin{equation}
0\le
\frac{m_{t,i}^{+}}{D_{t,i}^{+}}
+\frac{m_{t,i}^{-}}{D_{t,i}^{-}}
\le\frac{2bP_t}{A\sqrt{V_t}+\eps_t}.
\label{eq:paired-update-bound}
\end{equation}
\end{lemma}

The pair has a positive first-moment sum. The absolute difference between
its denominators, divided by their sum, is at most $b/(2A+b)$. This limits
the effect of unequal normalization on the sum of the two updates.
The proof is in Appendix~\ref{app:pair-proof}.

\paragraph{The pathwise invariant.}
On $\mathcal A$, suppose $Y(z_s)\ge1$ up to time $t$, so the observations equal $q_s$. The control moments are $m_{t,0}=-\gamma\delta P_t$ and $v_{t,0}=\gamma^2\delta^2V_t$. Since $V_t>0$ and each pair has $v_{t,i}^{\pm}\ge A^2V_t$, all denominators are positive even when $\eps_t=0$. The control contribution and the total contribution of the pairs satisfy
\begin{align*}
\gamma\delta(u_{t+1}-u_t)
&=\frac{\gamma^2\delta\eta_tP_t}{\gamma\sqrt{V_t}+\frac{\eps_t}{\delta}},\\
0\le-b\sum_i(s_{t+1,i}-s_{t,i})
&\le\frac{Nb^2\eta_tP_t}{A\sqrt{V_t}+\eps_t}
=\frac{\gamma^2\delta\eta_tP_t}{\gamma\sqrt{V_t}+N\eps_t}.
\end{align*}
Consequently,
\begin{equation}
Y(z_{t+1})-Y(z_t)
\ge\gamma^2\delta\eta_tP_t
\left[\frac1{\gamma\sqrt{V_t}+\frac{\eps_t}{\delta}}
-\frac1{\gamma\sqrt{V_t}+N\eps_t}\right]\ge0,
\label{eq:main-monotonicity}
\end{equation}
because $N\ge\delta^{-1}$. Starting from $Y(z_1)=1$, induction keeps every iterate in the affine region on $\mathcal A$. Hence $R_T\ge(7/8)\gamma\sqrt\delta$.

Appendix~\ref{app:lower} gives the full construction and induction,
and Appendix~\ref{app:lower-verification} verifies $L\le1$ and $\Sigma\le1$.
Choosing $\delta=T^{-\frac{p-1}{p}}$ for sufficiently large $T$
gives the stated time exponent with $d=1+2\lceil\delta^{-1}\rceil$.

\section{Conclusion}

Adam is usually viewed as a coordinatewise adaptive method because its second-moment estimate is coordinatewise. The results here show that this view is incomplete. The denominator contains two scales, $\sqrt{v_{t,i}}$ and $\eps$. When $\eps$ is shared, the normalization is only as coordinatewise as a single stabilization scale allows it to be. If the relevant noise and curvature scales differ substantially across coordinates, one scalar is asked to regularize several incompatible local regimes.

This points to a distinction that is easy to miss in adaptive methods. Coordinatewise statistics do not by themselves imply coordinatewise tuning. A global regularization applied after those statistics can reintroduce a common scale and limit the adaptivity that the coordinatewise estimator was meant to provide. For Adam, $\eps$ is a particularly simple instance of this mismatch.

\bibliography{adam_coordinate_calibration}
\bibliographystyle{plainnat}

\clearpage
\appendix

\section{Upper bound and balanced-scale corollary}
\label{app:upper}

The proof has two stages. First, we control the accumulated tracking error
and use smooth descent to bound $T^{-1}\sum_{t,i}\E[h_{t,i}^2/D_{t,i}]$.
Second, we bound the average denominator and apply Cauchy--Schwarz to
recover the unweighted $\ell_1$ criterion. Appendix~\ref{app:selfnorm}
proves the noise estimate used in the first stage. Appendix~\ref{app:corollaries} proves the balanced-scale corollary.

Throughout the appendices, $c$ and $C$ may change from line to line and may
depend on $p$ and $\mathcal K$. Additional dependence is stated when it first
appears. For brevity, write
\[
R_\star=(\Delta L)^{\frac{p-1}{3p-2}}
\Sigma^{\frac{p}{3p-2}}T^{-\frac{p-1}{3p-2}}.
\]
Also write
\[
h_t=\grad f(x_t),\qquad
D_{t,i}=\sqrt{v_{t,i}}+\eps_i.
\]
Set $x_0=x_1$, $h_0=h_1$, $e_0=-h_1$, and, for $t\ge1$,
\[
e_t=m_t-h_t.
\]
Subtracting $h_t$ from the first-moment update separates gradient drift
from momentum noise:
\begin{equation}
\begin{aligned}
e_t=m_t-h_t
&=\beta_1(m_{t-1}-h_{t-1})+\beta_1(h_{t-1}-h_t)+\rho_1\xi_t\\
&=\beta_1e_{t-1}+\beta_1(h_{t-1}-h_t)+\rho_1\xi_t.
\end{aligned}
\label{eq:tracking}
\end{equation}
Define
\begin{equation}
\kappa=\max_i\frac{L_i}{\eps_i}.
\label{eq:eps-kappa}
\end{equation}
The choice of $\eps_i$ gives
\begin{equation}
\begin{aligned}
\sum_i\eps_i
&\le\rho_2^{-\frac{2-p}{2p}}\left(\sum_i\sigma_i+
\frac{\Sigma}{L}\sum_iL_i\right)=2\Sigma\rho_2^{-\frac{2-p}{2p}},\\
\kappa
&\le\max_i\frac{L_i}{\rho_2^{-\frac{2-p}{2p}}(L_i/L)\Sigma}
=\frac{L}{\Sigma}\rho_2^{\frac{2-p}{2p}},\\
\eta\kappa
&\le\frac{\lambda\rho_1\Sigma\rho_2^{-\frac{2-p}{2p}}}{L}
\frac{L}{\Sigma}\rho_2^{\frac{2-p}{2p}}=\lambda\rho_1.
\end{aligned}
\label{eq:eps-kappa-bounds}
\end{equation}

All expectations below are finite on each fixed horizon. For
$0<\beta_2<1$, weighted Cauchy--Schwarz in the moment recursions gives
\[
\frac{|m_{t,i}|}{D_{t,i}}\le\frac{\rho_1}{\sqrt{\rho_2}}
\left(\sum_{k=0}^{t-1}\left(\frac{\beta_1^2}{\beta_2}\right)^k\right)^{\frac12}.
\]
Thus the displacements, true gradients, and objective values are bounded
on that horizon. Finite first moments of the noise give
$\E\sqrt{v_{t,i}}<\infty$; the same Cauchy--Schwarz bound and
$\eps_i>0$ imply integrability of $D_{t,i}$,
$m_{t,i}^2/D_{t,i}$, and $e_{t,i}^2/D_{t,i}$.

\subsection{Coordinatewise tracking}

The descent argument needs the weighted tracking error to be a small
fraction of the weighted gradient term, up to noise, initialization,
and drift terms. The following estimate provides this form by expanding
the momentum and splitting its noise at $\eps_i/\sqrt{\rho_2}$.

\begin{lemma}[Coordinatewise tracking estimate]
\label{lem:self-normalization}
For every $\tau\in(0,1)$, there exists $\bar\rho>0$, depending on
$p$, $\mathcal K$, and $\tau$, such that, whenever
$\rho_2/\rho_1\in\mathcal K$ and $0<\rho_1\le\bar\rho$,
\begin{align}
\sum_{t=1}^T\sum_i\E\frac{e_{t,i}^2}{D_{t,i}}
&\le
\tau\sum_{t=1}^T\sum_i\E\frac{h_{t,i}^2}{D_{t,i}}
+C\left(
\rho_1T\sum_i\eps_i+
\sum_i\frac{e_{0,i}^2}{\eps_i\delta_p}
\right)\nonumber\\
&\quad+
\frac{C}{\delta_p^2}
\sum_{t=1}^T\sum_i
\frac{\beta_1^2\E(h_{t-1,i}-h_{t,i})^2}{\eps_i}.
\label{eq:selfnorm-vector}
\end{align}
\end{lemma}

Appendix~\ref{app:selfnorm} derives the lemma from the conditional moment assumption. We next control the initial state and the gradient drift.

\subsection{Proof of Theorem~\ref{thm:vector-upper}}
\label{app:upper-proof}

\begin{proof}
We first control tracking, then bound the weighted gradient term, and
finally recover the unweighted stationarity criterion.

\paragraph{Tracking error.}
With $e_0=-h_1$ and $y=x_1-D_L^{-1}h_1$, smooth descent gives
\begin{align*}
\inf f\le f(y)
&\le f(x_1)-\|h_1\|_{D_L^{-1}}^2
+\frac12\|h_1\|_{D_L^{-1}}^2
=f(x_1)-\frac12\|h_1\|_{D_L^{-1}}^2,\\
\sum_i\frac{e_{0,i}^2}{\eps_i}
&\le\kappa\|h_1\|_{D_L^{-1}}^2
\le2\kappa(f(x_1)-\inf f)\le2\kappa\Delta.
\end{align*}
Coordinatewise smoothness, the Adam update, and $D_{t,i}\ge\eps_i$ give
\begin{align*}
\sum_{t=1}^T\sum_i
\E\frac{\beta_1^2(h_{t-1,i}-h_{t,i})^2}{\eps_i}
&\le
\beta_1^2\kappa\sum_{t=1}^T
\E\|h_{t-1}-h_t\|_{D_L^{-1}}^2\\
&\le
\beta_1^2\eta^2\kappa\sum_{t=1}^T
\E\left\|\left(\frac{m_{t-1,i}}{D_{t-1,i}}\right)_i\right\|_{D_L}^2\\
&\le\beta_1^2\eta^2\kappa\sum_{t=1}^T\sum_i
\E\left[\frac{L_i}{D_{t,i}}
\frac{(h_{t,i}+e_{t,i})^2}{D_{t,i}}\right]\\
&\le
2\beta_1^2\eta^2\kappa^2
\sum_{t=1}^T\sum_i
\E\frac{h_{t,i}^2+e_{t,i}^2}{D_{t,i}}.
\end{align*}
The shift of the update sum uses $m_0=0$.

The drift estimate reintroduces tracking error on the right. We make
its coefficient small using $\eta\kappa\le\lambda\rho_1$.
Uniformly over $\rho_2/\rho_1\in\mathcal K$, choose
$c_\delta>0$ with $\delta_p\ge c_\delta\rho_1$ for small $\rho_1$.
Fix $\tau=1/16$ and then $\lambda$ small enough that
$C\eta^2\kappa^2/\delta_p^2\le C\lambda^2/c_\delta^2\le1/16$.
Substitution into Lemma~\ref{lem:self-normalization} gives
\begin{align}
\sum_{t=1}^T\sum_i\E\frac{e_{t,i}^2}{D_{t,i}}
&\le\frac{\tau+C\eta^2\kappa^2/\delta_p^2}
{1-C\eta^2\kappa^2/\delta_p^2}
\sum_{t=1}^T\sum_i\E\frac{h_{t,i}^2}{D_{t,i}}\nonumber\\
&\quad+\frac{C}{1-C\eta^2\kappa^2/\delta_p^2}
\left(\rho_1T\sum_i\eps_i+\frac{\kappa\Delta}{\delta_p}\right)\nonumber\\
&\le\frac2{15}\sum_{t,i}\E\frac{h_{t,i}^2}{D_{t,i}}
+C\left(\rho_1T\sum_i\eps_i+\frac{\kappa\Delta}{\rho_1}\right)\nonumber\\
&\le\frac14\sum_{t=1}^T\sum_i\E\frac{h_{t,i}^2}{D_{t,i}}
+C\left(
\rho_1T\sum_i\eps_i+\frac{\kappa\Delta}{\rho_1}
\right).
\label{eq:tracking-aggregate}
\end{align}

\paragraph{Weighted descent.}
Young's inequality $|ab|\le a^2/4+b^2$ gives
$h_{t,i}(h_{t,i}+e_{t,i})\ge(3/4)h_{t,i}^2-e_{t,i}^2$.
Therefore \eqref{eq:tracking-aggregate} gives
\begin{align*}
\sum_{t=1}^T\sum_i\E\frac{h_{t,i}m_{t,i}}{D_{t,i}}
&=
\sum_{t=1}^T\sum_i
\E\frac{h_{t,i}(h_{t,i}+e_{t,i})}{D_{t,i}}\\
&\ge
\frac34\sum_{t=1}^T\sum_i
\E\frac{h_{t,i}^2}{D_{t,i}}
-\sum_{t=1}^T\sum_i
\E\frac{e_{t,i}^2}{D_{t,i}}\\
&\ge
\frac12\sum_{t=1}^T\sum_i
\E\frac{h_{t,i}^2}{D_{t,i}}
-C\left(
\rho_1T\sum_i\eps_i+\frac{\kappa\Delta}{\rho_1}
\right).
\end{align*}
Similarly,
\begin{align*}
\sum_{t=1}^T\sum_iL_i\E\frac{m_{t,i}^2}{D_{t,i}^2}
&\le
2\kappa\sum_{t=1}^T\sum_i
\E\frac{h_{t,i}^2+e_{t,i}^2}{D_{t,i}}\\
&\le
\frac52\kappa\sum_{t=1}^T\sum_i\E\frac{h_{t,i}^2}{D_{t,i}}
+C\kappa\left(\rho_1T\sum_i\eps_i+\frac{\kappa\Delta}{\rho_1}\right).
\end{align*}
Applying \eqref{eq:smooth-descent} to the Adam update and
summing over $t$ now gives
\begin{align*}
\Delta
&\ge
\sum_{t=1}^T\E[f(x_t)-f(x_{t+1})]\\
&\ge
\eta\sum_{t=1}^T\sum_i\E\frac{h_{t,i}m_{t,i}}{D_{t,i}}
-\frac{\eta^2}{2}\sum_{t=1}^T\sum_iL_i
\E\frac{m_{t,i}^2}{D_{t,i}^2}\\
&\ge
\eta\left(\frac12-\frac54\eta\kappa\right)
\sum_{t=1}^T\sum_i
\E\frac{h_{t,i}^2}{D_{t,i}}
-C\eta(1+\eta\kappa)\left(
\rho_1T\sum_i\eps_i+\frac{\kappa\Delta}{\rho_1}
\right).
\end{align*}
For $\lambda\le1/5$, we have $\eta\kappa\le1/5$, and therefore
\begin{equation}
\begin{aligned}
\frac1T\sum_{t=1}^T\sum_i
\E\frac{h_{t,i}^2}{D_{t,i}}
&\le\frac{4\Delta}{\eta T}
+C\rho_1\sum_i\eps_i+\frac{C\kappa\Delta}{\rho_1T}\\
&\le\frac{(4+C\lambda)\Delta}{\eta T}+C\rho_1\sum_i\eps_i
\le C\left(
\frac{\Delta}{\eta T}+\rho_1\sum_i\eps_i
\right).
\end{aligned}
\label{eq:adaptive-descent}
\end{equation}

Write
\[
S=\sum_i\eps_i.
\]
By \eqref{eq:coordinate-scales} and \eqref{eq:eps-kappa-bounds},
$S\asymp\Sigma\rho_2^{-\frac{2-p}{2p}}$,
$\eta\asymp\rho_1S/L$, and $\rho_1\asymp\rho_2\asymp\rho_\star$.
By the definition of $\rho_\star$,
\begin{equation}
\begin{aligned}
\rho_1^2S^2
&\asymp\Sigma^2\rho_\star^{\frac{3p-2}{p}}
=\frac{\Delta L}{T},\\
\frac{\Delta}{\eta T}
&\asymp\frac{\Delta L}{\rho_1ST}
\asymp\frac{\rho_1^2S^2}{\rho_1S}=\rho_1S,\\
\frac1T\sum_{t=1}^T\sum_i\E\frac{h_{t,i}^2}{D_{t,i}}
&\le C\left(\frac{\Delta}{\eta T}+\rho_1S\right)
\le C\rho_1S.
\end{aligned}
\label{eq:tuned-descent}
\end{equation}

\paragraph{Average denominator.}
We bound past gradient contributions using the weighted gradient term from
\eqref{eq:tuned-descent}. The recursion and $\eps_i>0$ give, for $s\le t$,
\begin{align*}
D_{t,i}&=\sqrt{v_{t,i}}+\eps_i
\ge\beta_2^{\frac{t-s}{2}}\sqrt{v_{s,i}}+\eps_i
\ge\beta_2^{\frac{t-s}{2}}D_{s,i},\\
\rho_2\sum_{s=1}^t\beta_2^{t-s}h_{s,i}^2
&=\rho_2\sum_{s=1}^t\beta_2^{t-s}D_{s,i}\frac{h_{s,i}^2}{D_{s,i}}\\
&\le D_{t,i}\rho_2\sum_{s=1}^t\beta_2^{\frac{t-s}{2}}
\frac{h_{s,i}^2}{D_{s,i}}.
\end{align*}
Using $v_{0,i}=0$, the weighted Euclidean triangle inequality, and
$2\sqrt{ab}\le a+b$, we obtain
\begin{align*}
D_{t,i}
&=\left(\rho_2\sum_{s=1}^t\beta_2^{t-s}
(h_{s,i}+\xi_{s,i})^2\right)^{1/2}+\eps_i\\
&\le\left(\rho_2\sum_{s=1}^t\beta_2^{t-s}h_{s,i}^2\right)^{1/2}
+\left(\rho_2\sum_{s=1}^t\beta_2^{t-s}\xi_{s,i}^2\right)^{1/2}+\eps_i\\
&\le\frac{D_{t,i}}2
+\frac{\rho_2}{2}\sum_{s=1}^t\beta_2^{\frac{t-s}{2}}
\frac{h_{s,i}^2}{D_{s,i}}
+\left(\rho_2\sum_{s=1}^t\beta_2^{t-s}\xi_{s,i}^2\right)^{1/2}+\eps_i.
\end{align*}
Consequently,
\begin{equation}
D_{t,i}\le\rho_2\sum_{s=1}^t\beta_2^{\frac{t-s}{2}}
\frac{h_{s,i}^2}{D_{s,i}}
+2\left(\rho_2\sum_{s=1}^t\beta_2^{t-s}\xi_{s,i}^2\right)^{1/2}
+2\eps_i.
\label{eq:denom-decomp}
\end{equation}
Exchanging the time sums gives
\begin{align}
\rho_2\sum_{t=1}^T\sum_{s=1}^t\beta_2^{\frac{t-s}{2}}
\frac{h_{s,i}^2}{D_{s,i}}
&=\rho_2\sum_{s=1}^T\frac{h_{s,i}^2}{D_{s,i}}
\sum_{k=0}^{T-s}\beta_2^{k/2}\nonumber\\
&\le\frac{\rho_2}{1-\sqrt{\beta_2}}
\sum_{s=1}^T\frac{h_{s,i}^2}{D_{s,i}}\nonumber\\
&=(1+\sqrt{\beta_2})\sum_{s=1}^T\frac{h_{s,i}^2}{D_{s,i}}
\le2\sum_{s=1}^T\frac{h_{s,i}^2}{D_{s,i}}.
\label{eq:denominator-energy}
\end{align}
For the noise term, subadditivity of $z\mapsto z^{\frac{p}{2}}$,
\eqref{eq:oracle}, and Lyapunov's inequality give
\begin{align*}
\E\left(\rho_2\sum_{s=1}^t\beta_2^{t-s}\xi_{s,i}^2\right)^{1/2}
&\le\left[\E\left(\rho_2\sum_{s=1}^t\beta_2^{t-s}\xi_{s,i}^2\right)^{p/2}\right]^{1/p}\\
&\le\left(\rho_2^{p/2}\sum_{s=1}^t\beta_2^{\frac{(t-s)p}{2}}
\E|\xi_{s,i}|^p\right)^{1/p}\\
&\le\left(\frac{\sigma_i^p\rho_2^{p/2}}{1-\beta_2^{p/2}}\right)^{1/p}
\le C\sigma_i\rho_2^{-\frac{2-p}{2p}}
\le C\eps_i.
\end{align*}
Here concavity gives
$1-\beta_2^{p/2}\ge(p/2)(1-\beta_2)=(p/2)\rho_2$.

Combining this noise bound with
\eqref{eq:denom-decomp}--\eqref{eq:denominator-energy} proves
\begin{equation}
\frac1T\sum_{t=1}^T\sum_i\E D_{t,i}
\le 2\left(\frac1T\sum_{t=1}^T\sum_i
\E\frac{h_{t,i}^2}{D_{t,i}}\right)+CS.
\label{eq:denominator-sum}
\end{equation}
\paragraph{Recovering stationarity.}
Write $|h_{t,i}|=(|h_{t,i}|/\sqrt{D_{t,i}})\sqrt{D_{t,i}}$.
Cauchy--Schwarz jointly over time, coordinates, and probability,
followed by \eqref{eq:denominator-sum} and \eqref{eq:tuned-descent}, gives
\begin{align*}
R_T^2
&\le
\left(\frac1T\sum_{t=1}^T\sum_i\E\frac{h_{t,i}^2}{D_{t,i}}\right)
\left(\frac1T\sum_{t=1}^T\sum_i\E D_{t,i}\right)\\
&\le
\left(\frac1T\sum_{t=1}^T\sum_i\E\frac{h_{t,i}^2}{D_{t,i}}\right)
\left(2\frac1T\sum_{t=1}^T\sum_i
\E\frac{h_{t,i}^2}{D_{t,i}}+CS\right)\\
&\le C\rho_1S(\rho_1S+S)\le C\rho_1S^2.
\end{align*}
Taking square roots gives
\begin{align*}
R_T&\le C\sqrt{\rho_1}\,S
\le C\Sigma\rho_\star^{\frac{p-1}{p}}\\
&=C\Sigma\left(\frac{\Delta L}{\Sigma^2T}\right)^{\frac{p-1}{3p-2}}
=C(\Delta L)^{\frac{p-1}{3p-2}}
\Sigma^{\frac{p}{3p-2}}T^{-\frac{p-1}{3p-2}}=CR_\star.
\end{align*}
This proves the theorem.
\end{proof}

\subsection{Proof of the coordinatewise tracking estimate}
\label{app:selfnorm}

We prove Lemma~\ref{lem:self-normalization} by expanding the momentum.
The noise decomposition controls small observations through their truncated
second moments. For large observations, we retain the random denominator and
sum over subsequent iterations using the recursions for $m_t$ and $v_t$.
Fix a coordinate and suppress its index. Write $h_t=\partial_i f(x_t)$,
$D_t=\sqrt{v_t}+\eps$, and $h_0=h_1$. The condition on $\eps$ is
\begin{equation}
\sigma^p\le\eps^p\rho_2^{\frac{2-p}{2}}.
\label{eq:tracking-floor-condition}
\end{equation}
All estimates below are uniform over $\rho_2/\rho_1\in\mathcal K$ for
sufficiently small $\rho_1$, so $\delta_p\asymp\rho_1$.
Put $r=\rho_2/\rho_1$. Taylor expansion
uniformly for $r\in\mathcal K$ gives
\begin{align*}
\frac{\beta_1^2}{\beta_2^{(2-p)/2}}
&=(1-2\rho_1+\rho_1^2)
\left(1+\frac{(2-p)r}{2}\rho_1+O(\rho_1^2)\right)\\
&=1-\left(2-\frac{(2-p)r}{2}\right)\rho_1+O(\rho_1^2).
\end{align*}
The coefficient $2-(2-p)r/2$ has a strictly positive minimum on $\mathcal K$;
at $p=2$ it equals $2$. Thus there are constants $c_\delta,C_\delta>0$
such that $c_\delta\rho_1\le\delta_p\le C_\delta\rho_1$ for sufficiently small $\rho_1$.

\subsubsection{The accumulated noise}

Let
\[
n_t=\rho_1\sum_{s=1}^t\beta_1^{t-s}\xi_s.
\]
We show that
\begin{equation}
\sum_{t=1}^T\E\frac{n_t^2}{D_t}
\le C\rho_1T\eps
+C\rho_1\rho_2\sum_{t=1}^T\E\frac{h_t^2}{D_t}.
\label{eq:momentum-noise-bound}
\end{equation}
At $p=2$, conditional centering makes distinct noise terms orthogonal.
Indeed, for $s<j$,
$\E(\xi_s\xi_j)=\E[\xi_s\E(\xi_j\mid\F_{j-1})]=0$.
We first remove the random denominator using $D_t\ge\eps$, and then
apply this orthogonality. Since $\sigma\le\eps$,
\[
\E\frac{n_t^2}{D_t}
\le\frac{\rho_1^2}{\eps}
\sum_{s=1}^t\beta_1^{2(t-s)}\E\xi_s^2
\le\frac{\rho_1^2\eps}{1-\beta_1^2}
\le\rho_1\eps.
\]
This proves \eqref{eq:momentum-noise-bound} at the endpoint. Assume
$1<p<2$ for the rest of this subsection.

\paragraph{Small and large noise.}
Set $B=\eps/\sqrt{\rho_2}$ and define
\[
a_t=\xi_t\mathbf1_{\{|\xi_t|\le B\}}
-\E[\xi_t\mathbf1_{\{|\xi_t|\le B\}}\mid\F_{t-1}],
\qquad b_t=\xi_t-a_t.
\]
To see both the centering and the moment bounds, write
\[
\mu_t=\E[\xi_t\mathbf1_{\{|\xi_t|\le B\}}\mid\F_{t-1}]
=-\E[\xi_t\mathbf1_{\{|\xi_t|>B\}}\mid\F_{t-1}].
\]
Then $b_t=\xi_t\mathbf1_{\{|\xi_t|>B\}}+\mu_t$, and both
$a_t$ and $b_t$ have conditional mean zero. Their moments satisfy
\begin{align}
\E[a_t^2\mid\F_{t-1}]
&\le\E[\xi_t^2\mathbf1_{\{|\xi_t|\le B\}}\mid\F_{t-1}]
\le B^{2-p}\sigma^p\le\eps^2,\nonumber\\
\E[|b_t|\mid\F_{t-1}]
&\le2\E[|\xi_t|\mathbf1_{\{|\xi_t|>B\}}\mid\F_{t-1}]
\le2B^{1-p}\sigma^p\le2\eps\sqrt{\rho_2}.
\label{eq:split-noise-moments}
\end{align}
For the small-noise sum, remove the denominator before using
orthogonality:
\begin{align*}
\E\frac{\bigl(\rho_1\sum_{s\le t}\beta_1^{t-s}a_s\bigr)^2}{D_t}
&\le\frac{\rho_1^2}{\eps}
\sum_{s\le t}\beta_1^{2(t-s)}\E a_s^2\\
&\le\frac{\rho_1^2\eps}{1-\beta_1^2}
\le\rho_1\eps.
\end{align*}
For $s<j$, $b_s$ is $\F_{j-1}$-measurable, so
\[
\E|b_sb_j|
 =\E\bigl[|b_s|\E(|b_j|\mid\F_{j-1})\bigr]
\le2\eps\sqrt{\rho_2}\,\E|b_s|
\le4\eps^2\rho_2.
\]
Using $D_t\ge\eps$, the cross terms in the large-noise square satisfy
\begin{equation}
\begin{aligned}
2\rho_1^2\sum_{s<j\le t}\beta_1^{2t-s-j}
\E\frac{|b_sb_j|}{D_t}
&\le4\eps\rho_2\left(2\rho_1^2
\sum_{s<j\le t}\beta_1^{t-s}\beta_1^{t-j}\right)\\
&\le4\eps\rho_2
\left(\rho_1\sum_{s\le t}\beta_1^{t-s}\right)^2
\le4\eps\rho_2.
\end{aligned}
\label{eq:large-noise-cross}
\end{equation}

\paragraph{A single diagonal term.}
Put $\bar D_t=\sqrt{\beta_2v_{t-1}}+\eps$ and
$J_t=\E[h_t^2/D_t\mid\F_{t-1}]$.
First we bound a single squared noise term by
\begin{equation}
\E\left[\frac{\xi_t^2}{\eps^{p-1}D_t^{2-p}}
\,\middle|\,\F_{t-1}\right]
\le C\eps\left[1+
\left(\frac{\sqrt{\rho_2}|h_t|}{\bar D_t}\right)^{2-p}\right].
\label{eq:fresh-square}
\end{equation}
Indeed, on $|\xi_t|>2|h_t|$, the bound
$D_t\ge\sqrt{\rho_2}|\xi_t|/2$ gives
\[
\frac{\xi_t^2}{\eps^{p-1}D_t^{2-p}}
\le\frac{C|\xi_t|^p}{\eps^{p-1}\rho_2^{\frac{2-p}{2}}}.
\]
On the complement, use
$|\xi_t|^2\le(2|h_t|)^{2-p}|\xi_t|^p$ and $D_t\ge \bar D_t$.
Both $h_t$ and $\bar D_t$ are $\F_{t-1}$-measurable. Thus the conditional
contributions of the two events are bounded, respectively, by
\[
\frac{C\sigma^p}{\eps^{p-1}\rho_2^{(2-p)/2}}\le C\eps,
\qquad
\frac{(2|h_t|)^{2-p}\sigma^p}{\eps^{p-1}\bar D_t^{2-p}}
\le C\eps\left(\frac{\sqrt{\rho_2}|h_t|}{\bar D_t}\right)^{2-p}.
\]
Adding these bounds proves \eqref{eq:fresh-square}.

To control its gradient-dependent term, use
$D_t\le \bar D_t+\sqrt{\rho_2}|h_t+\xi_t|$ together with the conditional
first-moment bound $\E[|\xi_t|\mid\F_{t-1}]\le\sigma$ to obtain
\begin{align*}
\E[D_t\mid\F_{t-1}]
&\le \bar D_t+\sqrt{\rho_2}(|h_t|+\sigma)\\
&\le2\bar D_t+\sqrt{\rho_2}|h_t|.
\end{align*}
Here the condition on $\eps$ implies
$\sqrt{\rho_2}\sigma\le\eps\rho_2^{1/p}\le\eps\le \bar D_t$.
Conditional Jensen and the preceding denominator bound give
\begin{equation}
\begin{aligned}
J_t&=h_t^2\E[D_t^{-1}\mid\F_{t-1}]
\ge\frac{h_t^2}{\E[D_t\mid\F_{t-1}]}\\
&\ge\frac{h_t^2}{2\bar D_t+\sqrt{\rho_2}|h_t|}
=\frac{\bar D_t}{\rho_2}
\frac{(\sqrt{\rho_2}|h_t|/\bar D_t)^2}
{2+\sqrt{\rho_2}|h_t|/\bar D_t}\\
&\ge\frac{\bar D_t}{2\rho_2}
\frac{(\sqrt{\rho_2}|h_t|/\bar D_t)^2}
{1+\sqrt{\rho_2}|h_t|/\bar D_t}.
\end{aligned}
\label{eq:normalized-gradient-lower}
\end{equation}
For $1<p<2$ and every $x\ge0$,
\[
x^{2-p}\le1+\frac{2x^2}{1+x}:
\]
the constant term suffices for $x\le1$, and for $x\ge1$ use
$x^{2-p}\le x\le2x^2/(1+x)$. Applying this inequality with
$x=\sqrt{\rho_2}|h_t|/\bar D_t$ and using $\eps\le\bar D_t$ gives
\[
\eps\left(\frac{\sqrt{\rho_2}|h_t|}{\bar D_t}\right)^{2-p}
\le\eps+
\frac{2\eps(\sqrt{\rho_2}|h_t|/\bar D_t)^2}
{1+\sqrt{\rho_2}|h_t|/\bar D_t}
\le\eps+
\frac{2\bar D_t(\sqrt{\rho_2}|h_t|/\bar D_t)^2}
{1+\sqrt{\rho_2}|h_t|/\bar D_t}
\le\eps+4\rho_2J_t.
\]
In the decomposition of $b_t$, conditional centering gives
$|\mu_t|\le B^{1-p}\sigma^p\le\eps\sqrt{\rho_2}$.
The centering correction contributes at most
\[
\frac{\mu_t^2}{\eps^{p-1}D_t^{2-p}}
\le\frac{\mu_t^2}{\eps}\le\eps\rho_2.
\]
Combining $b_t^2\le2\xi_t^2+2\mu_t^2$ with
\eqref{eq:fresh-square} and the preceding bounds gives
\begin{equation}
\begin{aligned}
\E\left[\frac{b_t^2}{\eps^{p-1}D_t^{2-p}}
\,\middle|\,\F_{t-1}\right]
&\le2\E\left[\frac{\xi_t^2}{\eps^{p-1}D_t^{2-p}}
\,\middle|\,\F_{t-1}\right]+\frac{2\mu_t^2}{\eps}\\
&\le C\eps\left[1+
\left(\frac{\sqrt{\rho_2}|h_t|}{\bar D_t}\right)^{2-p}\right]
+2\eps\rho_2
\le C\eps+C\rho_2J_t.
\end{aligned}
\label{eq:large-noise-diagonal}
\end{equation}

\paragraph{Summing over time.}
For $t\ge s$, the second-moment recursion gives
$D_t\ge\beta_2^{\frac{t-s}{2}}D_s$. Together with $D_t\ge\eps$,
this yields
\begin{align*}
\frac{b_s^2}{D_t}
&=\frac{b_s^2}{D_t^{p-1}D_t^{2-p}}
\le\frac{b_s^2}{\eps^{p-1}D_t^{2-p}}\\
&\le\beta_2^{-\frac{(t-s)(2-p)}{2}}
\frac{b_s^2}{\eps^{p-1}D_s^{2-p}}.
\end{align*}
Using $\beta_1^2/\beta_2^{(2-p)/2}=1-\delta_p$ and
$\E J_s=\E[h_s^2/D_s]$, summation gives
\begin{align*}
\rho_1^2\sum_{t=1}^T\sum_{s\le t}\beta_1^{2(t-s)}
\E\frac{b_s^2}{D_t}
&\le\rho_1^2\sum_{s=1}^T\sum_{k=0}^{T-s}
\left(\frac{\beta_1^2}{\beta_2^{\frac{2-p}{2}}}\right)^k
\E\frac{b_s^2}{\eps^{p-1}D_s^{2-p}}\\
&\le\frac{C\rho_1^2}{\delta_p}
\left(T\eps+\rho_2\sum_{s=1}^T\E\frac{h_s^2}{D_s}\right).
\end{align*}
To combine the estimates, write $n_t$ as its small- and large-noise
sums and use $(x+y)^2\le2x^2+2y^2$. Within the large-noise square,
bound the cross terms in absolute value by \eqref{eq:large-noise-cross}.
This gives
\begin{align*}
\sum_{t=1}^T\E\frac{n_t^2}{D_t}
&\le C(\rho_1+\rho_2)T\eps
+\frac{C\rho_1^2}{\delta_p}
\left(T\eps+\rho_2\sum_{s=1}^T\E\frac{h_s^2}{D_s}\right)\\
&\le C\rho_1T\eps
+C\rho_1\rho_2\sum_{s=1}^T\E\frac{h_s^2}{D_s}.
\end{align*}
This proves \eqref{eq:momentum-noise-bound}.

\subsubsection{Initialization and gradient drift}

Expanding \eqref{eq:tracking} gives
\[
e_t=\beta_1^te_0+
\sum_{s=1}^t\beta_1^{t-s+1}(h_{s-1}-h_s)+n_t.
\]
The initial contribution obeys
\[
\sum_{t=1}^T\E\frac{\beta_1^{2t}e_0^2}{D_t}
\le\frac{e_0^2}{\eps}\sum_{t=1}^T\beta_1^{2t}
\le\frac{e_0^2}{\eps}\frac{\beta_1^2}{1-\beta_1^2}
\le\frac{e_0^2}{\eps\rho_1}.
\]
For the drift, put $d_s=h_{s-1}-h_s$. Using $D_t\ge\eps$ and
weighted Cauchy--Schwarz gives
\begin{align*}
\sum_{t=1}^T\E
\frac{\bigl(\sum_{s\le t}\beta_1^{t-s+1}d_s\bigr)^2}{D_t}
&\le\frac{\beta_1^2}{\eps}\sum_{t=1}^T
\left(\sum_{s\le t}\beta_1^{t-s}\right)
\left(\sum_{s\le t}\beta_1^{t-s}\E d_s^2\right)\\
&\le\frac{\beta_1^2}{\eps\rho_1}
\sum_{t=1}^T\sum_{s\le t}\beta_1^{t-s}
\E d_s^2\\
&=\frac{\beta_1^2}{\eps\rho_1}
\sum_{s=1}^T\E d_s^2\sum_{t=s}^T\beta_1^{t-s}\\
&\le\frac{\beta_1^2}{\eps\rho_1^2}
\sum_{s=1}^T\E(h_{s-1}-h_s)^2.
\end{align*}
Apply $(x+y+z)^2\le3(x^2+y^2+z^2)$ to the three terms in $e_t$.
The preceding estimates and \eqref{eq:momentum-noise-bound} yield
\[
\sum_{t=1}^T\E\frac{e_t^2}{D_t}
\le C\rho_1\rho_2\sum_{t=1}^T\E\frac{h_t^2}{D_t}
+C\rho_1T\eps+\frac{Ce_0^2}{\eps\rho_1}
+\frac{C\beta_1^2}{\eps\rho_1^2}
\sum_{t=1}^T\E(h_{t-1}-h_t)^2.
\]
Choose $\rho_1$ small enough that $C\rho_1\rho_2\le\tau$.
This is possible uniformly over $\rho_2/\rho_1\in\mathcal K$.
Using $\delta_p\asymp\rho_1$ and summing over coordinates proves
\eqref{eq:selfnorm-vector}.

\subsection{Shared \texorpdfstring{$\eps$}{epsilon}: balanced scales and fixed dimension}
\label{app:corollaries}

For Corollary~\ref{thm:balanced}, \eqref{eq:balance} and the common choice of $\eps_i$ give
\[
\eps_i\ge\rho_2^{-\frac{2-p}{2p}}
\max\left\{\sigma_i,\frac{L_i}{L}\Sigma\right\},
\qquad
\sum_i\eps_i=C_0\rho_2^{-\frac{2-p}{2p}}\Sigma.
\]
These are the two bounds on $\eps_i$ used in
\eqref{eq:eps-kappa-bounds}. In particular,
$\kappa\le(L/\Sigma)\rho_2^{\frac{2-p}{2p}}$ and
$\eta\kappa\le\lambda\rho_1$ still hold. Thus the descent estimate and
the denominator estimate in Appendix~\ref{app:upper-proof} give
\[
\frac1T\sum_{t=1}^T\sum_i\E\frac{h_{t,i}^2}{D_{t,i}}
\le C\left(\frac{\Delta}{\eta T}+\rho_1S\right),
\qquad S=C_0\Sigma\rho_2^{-\frac{2-p}{2p}},
\]
and
\[
R_T^2\le
\left(\frac1T\sum_{t=1}^T\sum_i\E\frac{h_{t,i}^2}{D_{t,i}}\right)
\left(2\frac1T\sum_{t=1}^T\sum_i
\E\frac{h_{t,i}^2}{D_{t,i}}+CS\right).
\]
Summing the balance inequalities shows $C_0\ge1$. The same prescribed
$\eta$ and $\rho_1=\rho_\star$ satisfy
$\Delta/(\eta T)\lesssim\rho_1\Sigma\rho_2^{-\frac{2-p}{2p}}\le\rho_1S$.
Consequently,
\begin{align*}
R_T^2&\le C\rho_1S(\rho_1S+S)\le C\rho_1S^2,\\
R_T&\lesssim\sqrt{\rho_1}S
=C_0\Sigma\rho_1^{1/2}\rho_2^{-\frac{2-p}{2p}}
\asymp C_0\Sigma\rho_\star^{(p-1)/p}=C_0R_\star,
\end{align*}
as claimed.

\paragraph{Fixed dimension.}
For arbitrary coordinate scales, \eqref{eq:balance} holds with $C_0=d$,
since $\max_i\sigma_i\le\Sigma$ and $\max_iL_i\le L$.
Corollary~\ref{thm:balanced} therefore applies with the shared choice
$\eps=\Sigma\rho_2^{-\frac{2-p}{2p}}$ and the remaining parameters
of Theorem~\ref{thm:vector-upper}. The preceding estimate is linear in
$C_0$, so setting $C_0=d$ gives
\[
R_T\le Cd
(\Delta L)^{\frac{p-1}{3p-2}}
\Sigma^{\frac{p}{3p-2}}T^{-\frac{p-1}{3p-2}}
\]
whenever $\rho_\star\le c$, with $C$ depending only on $p$ and
$\mathcal K$. At every fixed dimension, a shared $\eps$ thus attains
the benchmark time exponent uniformly over coordinate scales,
with a factor linear in $d$.

\section{Proof of the lower bound for a shared \texorpdfstring{$\eps$}{epsilon}}
\label{app:lower}

We give the objective, oracle, and pathwise invariant for
Theorem~\ref{thm:scalar-impossibility}.

\subsection{Parameters and objective}

Fix $0<\delta\le\min\{\delta_0,T^{-\frac{p-1}{p}}\}$ and set
\begin{equation}
\gamma=\frac1{128},
\qquad
N=\lceil\delta^{-1}\rceil,
\qquad
A=\frac{\gamma}{N},
\qquad
b=\frac{\gamma\sqrt\delta}{N}.
\label{eq:lower-parameters}
\end{equation}

Let
$z=(u,w_{1,+},w_{1,-},\ldots,w_{N,+},w_{N,-})$,
write
\[
s_i=\frac{w_{i,+}+w_{i,-}}{2},
\]
and define
\begin{equation}
Y(z)=1+\gamma\delta\,u+b\sum_{i=1}^N s_i.
\label{eq:affine-argument}
\end{equation}
Let
\begin{equation}
\phi(y)=
\begin{cases}
0,&y\le0,\\
\frac{y^2}{2},&0<y<1,\\
y-\frac{1}{2},&y\ge1,
\end{cases}
\qquad
f(z)=\phi(Y(z)).
\label{eq:lower-objective}
\end{equation}
The function $\phi$ is convex and continuously differentiable, with
$\phi'(y)=\min\{1,\max\{0,y\}\}$. Since $\phi\ge0$ and $Y(z)$
can take any real value, $\inf f=0$. At $z_1=0$, $Y(z_1)=1$, so
\begin{equation}
f(z_1)-\inf f=\frac12.
\label{eq:lower-gap}
\end{equation}
Let $\varrho$ be the coefficient vector in \eqref{eq:affine-argument},
with $\varrho_0=\gamma\delta$ and
$\varrho_{i,+}=\varrho_{i,-}=b/2$.
The chain rule gives $\grad f(z)=\phi'(Y(z))\varrho$.
Whenever $Y(z)\ge1$,
\begin{equation}
\grad f(z)=\varrho,
\qquad
\|\grad f(z)\|_1=\gamma\delta+Nb=\gamma\delta+\gamma\sqrt\delta
\ge\gamma\sqrt\delta.
\label{eq:active-gradient}
\end{equation}

\subsection{Sample-function construction}

Let $\theta=1/(8T)$.  Independently over time, define a random vector $q_t$ as follows.  Its control coordinate is
\begin{equation}
q_{t,0}=
\begin{cases}
-\gamma\delta,&\text{with probability }1-\theta,\\
\gamma\delta(\frac{2}{\theta}-1),&\text{with probability }\theta.
\end{cases}
\label{eq:control-oracle}
\end{equation}
Draw one sign $\zeta_t\in\{-1,1\}$ with equal probabilities, independently
of the control coordinate, and use it for all pairs:
\begin{equation}
(q_{t,i}^{+},q_{t,i}^{-})=
\begin{cases}
(A+b,-A),&\zeta_t=1,\\
(-A,A+b),&\zeta_t=-1.
\end{cases}
\label{eq:pair-oracle}
\end{equation}
The two coordinates therefore receive a uniformly random permutation of
$\{A+b,-A\}$. Their means satisfy
\begin{equation}
\begin{aligned}
\E q_{t,0}&=(1-\theta)(-\gamma\delta)
+\theta\gamma\delta\left(\frac2\theta-1\right)=\gamma\delta,\\
\E q_{t,i}^{\pm}&=\frac{(A+b)+(-A)}2=\frac b2,
\qquad\text{hence }\E q_t=\varrho.
\end{aligned}
\label{eq:q-mean}
\end{equation}

Put $\omega_t=q_t-\varrho$ and define the random sample function
\begin{equation}
F_t(z)=f(z)+\langle\omega_t,z\rangle.
\label{eq:sample-function}
\end{equation}
Every $F_t$ is convex and continuously differentiable, and its gradient
\begin{equation}
g_t(z)=\phi'(Y(z))\varrho+\omega_t
\label{eq:lower-gradient-oracle}
\end{equation}
satisfies $\E[g_t(z)]=\grad f(z)$.  The noise $\omega_t$ is independent of $z$ and of the past.  Appendix~\ref{app:lower-verification} verifies its coordinatewise $p$th moments.

Define the event
\begin{equation}
\mathcal A=\{q_{t,0}=-\gamma\delta\text{ for every }1\le t\le T\}.
\label{eq:common-control-event}
\end{equation}
The union bound gives
\begin{equation}
\Pr(\mathcal A)=(1-\theta)^T\ge1-T\theta=\frac78.
\label{eq:control-event-prob}
\end{equation}

\subsection{Averaging coefficients for a realized history}

For the time-varying $\beta_{1,t},\beta_{2,t}$ in \eqref{eq:adam-family}, define, for $1\le s\le t$,
\begin{align}
\alpha_{t,s}
&=(1-\beta_{1,s})\prod_{r=s+1}^t\beta_{1,r},
&
\nu_{t,s}
&=(1-\beta_{2,s})\prod_{r=s+1}^t\beta_{2,r},
\label{eq:time-varying-kernels}\\
P_t&=\sum_{s=1}^t\alpha_{t,s},
&
V_t&=\sum_{s=1}^t\nu_{t,s}.
\end{align}
These coefficients are nonnegative.
Iterating the two moment recursions from $m_0=v_0=0$ gives
\[
m_{t,j}=\sum_{s=1}^t\alpha_{t,s}g_{s,j},\qquad
v_{t,j}=\sum_{s=1}^t\nu_{t,s}g_{s,j}^2.
\]
In particular, $\alpha_{t,t}=1-\beta_{1,t}>0$ and
$\nu_{t,t}=1-\beta_{2,t}>0$, so $P_t,V_t>0$.

Suppose $Y(z_s)\ge1$ for $s\le t$.  Then \eqref{eq:lower-gradient-oracle} equals $q_s$ at those iterates.  On $\mathcal A$, the control coordinate therefore satisfies
\begin{equation}
m_{t,0}=-\gamma\delta P_t,
\qquad
v_{t,0}=\gamma^2\delta^2V_t.
\label{eq:control-moments}
\end{equation}
The resulting contribution to $Y$ is
\begin{equation}
\gamma\delta(u_{t+1}-u_t)
=\frac{\eta_t\gamma^2\delta^2P_t}
{\gamma\delta\sqrt{V_t}+\eps_t}
=\frac{\gamma^2\delta\eta_tP_t}
{\gamma\sqrt{V_t}+\frac{\eps_t}{\delta}}.
\label{eq:control-displacement}
\end{equation}

\subsection{Proof of the paired-update lemma}
\label{app:pair-proof}

\begin{proof}[Proof of Lemma~\ref{lem:paired-updates}]
Suppress the pair and time indices. Since each observation pair sums to
$b$ and its difference has magnitude $2A+b$,
\[
m^++m^-=bP,\qquad |m^+-m^-|\le(2A+b)P.
\]
Each squared observation lies between $A^2$ and $(A+b)^2$, so
\[
A\sqrt V+\eps_t\le D^\pm\le(A+b)\sqrt V+\eps_t.
\]
Since $A,V>0$, both denominators are positive even when $\eps_t=0$.
If $\ell\le d_2\le d_1\le u$ and $\ell>0$, then
$(d_1-d_2)/(d_1+d_2)\le(u-\ell)/(u+\ell)$: the ratio increases with
$d_1$ and decreases with $d_2$. Applying this to the preceding interval,
\begin{equation}
\frac{|(D^+)^{-1}-(D^-)^{-1}|}{(D^+)^{-1}+(D^-)^{-1}}
=\frac{|D^+-D^-|}{D^++D^-}
\le\frac{b\sqrt V}{(2A+b)\sqrt V+2\eps_t}
\le\frac{b}{2A+b}.
\label{eq:pair-denominator-difference}
\end{equation}
The potentially negative term in the normalized pair sum satisfies
\begin{align*}
\left|\frac{m^+-m^-}{2}
\left(\frac1{D^+}-\frac1{D^-}\right)\right|
&\le\frac{(2A+b)P}{2}\frac{b}{2A+b}
\left(\frac1{D^+}+\frac1{D^-}\right)\\
&=\frac{bP}{2}\left(\frac1{D^+}+\frac1{D^-}\right).
\end{align*}
Consequently,
\begin{align*}
0\le{}&
\frac{m^+}{D^+}+\frac{m^-}{D^-}
\\
={}&\frac{bP}{2}\left(\frac1{D^+}+\frac1{D^-}\right)
+\frac{m^+-m^-}{2}\left(\frac1{D^+}-\frac1{D^-}\right)
\\
\le{}& bP\left(\frac1{D^+}+\frac1{D^-}\right)
\le\frac{2bP}{A\sqrt V+\eps_t}.
\end{align*}
\end{proof}

\subsection{The iterates remain in the affine region}

On $\mathcal A$, suppose $Y(z_s)\ge1$ for $s\le t$. The mean of pair $i$
changes by
\[
s_{t+1,i}-s_{t,i}
=-\frac{\eta_t}{2}\left(
\frac{m_{t,i}^{+}}{D_{t,i}^{+}}+
\frac{m_{t,i}^{-}}{D_{t,i}^{-}}\right).
\]
Lemma~\ref{lem:paired-updates} bounds their total contribution by
\begin{equation}
0\le-b\sum_{i=1}^N(s_{t+1,i}-s_{t,i})
\le\frac{Nb^2\eta_tP_t}{A\sqrt{V_t}+\eps_t}
=\frac{\gamma^2\delta\eta_tP_t}{\gamma\sqrt{V_t}+N\eps_t}.
\label{eq:pair-mean-step}
\end{equation}
Combining this with \eqref{eq:control-displacement} gives
\begin{equation}
\begin{aligned}
Y(z_{t+1})-Y(z_t)
&=\gamma\delta(u_{t+1}-u_t)
+b\sum_{i=1}^N(s_{t+1,i}-s_{t,i})\\
&\ge\gamma^2\delta\eta_tP_t
\left[
\frac1{\gamma\sqrt{V_t}+\frac{\eps_t}{\delta}}
-\frac1{\gamma\sqrt{V_t}+N\eps_t}
\right]\ge0.
\end{aligned}
\label{eq:affine-region-preserved}
\end{equation}
The bracket is nonnegative because $N\ge\delta^{-1}$.

At $z_1=0$, $Y(z_1)=1$.  Suppose inductively that $Y(z_s)\ge1$ for $s\le t$.  The preceding calculation gives $Y(z_{t+1})\ge Y(z_t)\ge1$.  On $\mathcal A$,
\begin{equation}
Y(z_t)\ge1
\quad\text{and}\quad
\|\grad f(z_t)\|_1\ge\gamma\sqrt\delta
\qquad(1\le t\le T).
\label{eq:lower-active-all-time}
\end{equation}
Using \eqref{eq:control-event-prob},
\begin{equation}
\frac1T\sum_{t=1}^T\E\|\grad f(z_t)\|_1
\ge\Pr(\mathcal A)\gamma\sqrt\delta
\ge\frac78\frac{\sqrt\delta}{128}.
\label{eq:lower-final}
\end{equation}
To complete the proof, we verify Assumptions~\ref{ass:gc} and~\ref{ass:oracle} with $L\le1$ and $\Sigma\le1$ in the next subsection.

\subsection{Verification of the lower-bound assumptions}
\label{app:lower-verification}

We verify the two quantitative assumptions in Theorem~\ref{thm:scalar-impossibility}.  Equation \eqref{eq:lower-gap} gives the initial-gap bound.

\subsubsection{Coordinatewise smoothness}

Set
\begin{equation}
L_j^\star=\|\varrho\|_1\varrho_j.
\label{eq:lower-L-certificate}
\end{equation}
All coordinates of $\varrho$ are positive. Substituting the definition
of $L_j^\star$ gives
\begin{equation}
\sum_j\frac{\varrho_j^2}{L_j^\star}
=\frac1{\|\varrho\|_1}\sum_j\varrho_j=1,
\qquad
\sum_jL_j^\star=\|\varrho\|_1\sum_j\varrho_j=\|\varrho\|_1^2.
\label{eq:lower-L-identities}
\end{equation}
Because $\phi'$ in \eqref{eq:lower-objective} is $1$-Lipschitz,
\begin{align}
\|\grad f(z')-\grad f(z)\|_{(D_L^\star)^{-1}}
&=|\phi'(Y(z'))-\phi'(Y(z))|
\|\varrho\|_{(D_L^\star)^{-1}}\\
&\le|\langle\varrho,z'-z\rangle|\\
&\le\|\varrho\|_{(D_L^\star)^{-1}}
\|z'-z\|_{D_L^\star}\\
&=\|z'-z\|_{D_L^\star}.
\label{eq:lower-GC-proof}
\end{align}
If
$\delta_0\le1$, then $\|\varrho\|_1=\gamma(\delta+\sqrt\delta)\le2\gamma=1/64$.
Thus $L^\star$ satisfies Assumption~\ref{ass:gc} and $\|L^\star\|_1\le1$.

\subsubsection{Coordinatewise moment bounds}

The oracle noise is $\omega_t=q_t-\varrho$. In the control coordinate it
equals $-2\gamma\delta$ on the common outcome and
$2\gamma\delta(1/\theta-1)$ on the rare outcome. Therefore
\begin{align}
\E|\omega_{t,0}|^p
&=(2\gamma\delta)^p\bigl[(1-\theta)+\theta(\frac{1}{\theta}-1)^p\bigr]\nonumber\\
&=(2\gamma\delta)^p\bigl[(1-\theta)+\theta^{1-p}(1-\theta)^p\bigr]\nonumber\\
&\le(2\gamma\delta)^p(1+\theta^{1-p}).
\label{eq:control-p-moment}
\end{align}
Choosing $\sigma_0=(\E|\omega_{t,0}|^p)^{1/p}$, we obtain
\begin{align}
\sigma_0
&\le2\gamma\delta(1+\theta^{1-p})^{1/p}
\le2\gamma\delta(1+\theta^{-\frac{p-1}{p}})\nonumber\\
&=\frac{\delta}{64}\left(1+(8T)^{\frac{p-1}{p}}\right)
\label{eq:control-sigma}\\
&\le\frac{1+8^{(p-1)/p}}{64}
\le\frac9{64}.
\label{eq:control-sigma-bound}
\end{align}
Here $\delta\le1$ and $\delta T^{(p-1)/p}\le1$ by the choice of $\delta$.

In pair $i$, each coordinate has mean $b/2$. Its two centered outcomes are
$(A+b)-b/2=A+b/2$ and $-A-b/2=-(A+b/2)$, each with probability $1/2$.
Thus its $p$th absolute moment is exactly $(A+b/2)^p$. We may take
$\sigma_{i,+}=\sigma_{i,-}=A+b/2$. From \eqref{eq:lower-parameters},
\begin{align}
\sum_{i=1}^N(\sigma_{i,+}+\sigma_{i,-})
&=2NA+Nb\\
&=\gamma(2+\sqrt\delta)
\le\frac3{128}.
\label{eq:pair-noise-total}
\end{align}
The sampling at time $t$ is independent
of $\F_{t-1}$, so these unconditional moment calculations also hold
conditionally, as required by Assumption~\ref{ass:oracle}.
Combining \eqref{eq:control-sigma-bound} and
\eqref{eq:pair-noise-total} gives
\begin{equation}
\Sigma
=\sigma_0+\sum_{i=1}^N(\sigma_{i,+}+\sigma_{i,-})
\le\frac9{64}+\frac3{128}=\frac{21}{128}<1.
\label{eq:lower-noise-budget}
\end{equation}
This completes the proof of Theorem~\ref{thm:scalar-impossibility}.

\section{One-dimensional optimality for fixed-parameter Adam}
\label{app:one-d}

When $d=1$, shared and coordinatewise choices of $\eps$ coincide. We record
the corresponding family-specific minimax statement as a supplementary result.

\begin{proposition}[Minimax rate within a fixed-parameter Adam family]
\label{prop:one-d}
Fix $p\in(1,2]$ and let $\mathcal K$ be a compact subset of
$(0,4/(2-p))$ when $p<2$, or of $(0,\infty)$ when $p=2$.
There exist $c,C,\bar\rho,\bar\lambda>0$, depending only on $p$ and
$\mathcal K$, such that, for each fixed ratio
$r=\rho_2/\rho_1\in\mathcal K$, the admissible parameters satisfy
\begin{equation}
\begin{gathered}
\rho_2=r\rho_1\le\tfrac12,\qquad
0<\rho_1\le\bar\rho,\qquad 0<\lambda\le\bar\lambda,\\
\eps=\Sigma\rho_2^{-\frac{2-p}{2p}},\qquad
\eta=\frac{\lambda\rho_1\Sigma\rho_2^{-\frac{2-p}{2p}}}{L}
=\frac{\lambda\rho_1\eps}{L}.
\end{gathered}
\label{eq:one-d-parameter-set}
\end{equation}
Whenever $\rho_\star\le c$ and $T\rho_\star\ge C$,
\[
\inf_{\rho_1,\lambda}
\sup_{f,O}
\frac1T\sum_{t=1}^T\E|f'(x_t)|
\asymp
(\Delta L)^{\frac{p-1}{3p-2}}
\Sigma^{\frac{p}{3p-2}}T^{-\frac{p-1}{3p-2}}.
\]
The infimum is over $(\rho_1,\lambda)$ satisfying
\eqref{eq:one-d-parameter-set}, and the supremum is over scalar
objective--oracle pairs satisfying
Assumptions~\ref{ass:gc}--\ref{ass:oracle}. The comparison is uniform over
$r\in\mathcal K$.
\end{proposition}

In one dimension, \eqref{eq:coordinate-scales} gives
$\eps=\Sigma\rho_2^{-\frac{2-p}{2p}}$. The upper half of
Proposition~\ref{prop:one-d} follows directly from
Theorem~\ref{thm:vector-upper}.

For the matching lower bound, we use sparse observations and measure each
nonzero observation by the total displacement it induces. Small displacements
limit travel, while large displacements permit a two-sign comparison with
identical denominators.

\subsection{Lower-bound construction}

Fix $r\in\mathcal K$ and a pair $(\rho_1,\lambda)$ satisfying
\eqref{eq:one-d-parameter-set}. We show that every such pair has worst-case
stationarity at least a constant multiple of $R_\star$.

Choose fixed small constants $\alpha,\mu>0$ and set
\begin{equation}
\theta=\alpha\rho_\star,
\qquad
A=\mu\Sigma\theta^{-\frac{1}{p}},
\qquad
H=\theta A.
\label{eq:scalar-lower-scales}
\end{equation}
Since $R_\star=\Sigma\rho_\star^{(p-1)/p}$,
\[
H=\theta A=\mu\Sigma\theta^{\frac{p-1}{p}}
=\mu\alpha^{\frac{p-1}{p}}R_\star,\qquad
T\rho_\star R_\star^2
=T\Sigma^2\rho_\star^{\frac{3p-2}{p}}=\Delta L.
\]
Fix $\alpha,\mu$ small enough that $\theta\le1/2$ and $\mu\le1/2$
under \eqref{eq:one-d-parameter-set} and $\rho_\star\le c$.
Define
\begin{equation}
\psi(y)=\clip\left(y,-\frac{1}{2},\frac{1}{2}\right),
\qquad
f'(z)=H\psi\left(\frac{Lz}{H}\right).
\label{eq:scalar-lower-objective}
\end{equation}
This derivative is nondecreasing and $L$-Lipschitz, so $f$ is convex and
$L$-smooth: $|f'(z)-f'(z')|\le H|Lz/H-Lz'/H|=L|z-z'|$.
Integrating and normalizing $f(0)=0$ gives
\[
f(z)=\begin{cases}
Lz^2/2,& |z|\le H/(2L),\\
H|z|/2-H^2/(8L),& |z|\ge H/(2L).
\end{cases}
\]
Thus $\inf f=f(0)=0$. Start from
\begin{equation}
z_1=\frac{H}{2L}+W,
\qquad
W=\frac{\Delta}{H}.
\label{eq:scalar-start}
\end{equation}
Since $f'(z)=H/2$ for $z\ge H/(2L)$,
\begin{equation}
\begin{aligned}
f(z_1)-\inf f
&=\frac{H^2}{8L}+\frac{HW}{2}
=\frac{H^2}{8L}+\frac\Delta2\\
&=\frac{\mu^2\alpha^{2(p-1)/p}}{8T\rho_\star}\Delta
+\frac\Delta2\le\Delta.
\end{aligned}
\label{eq:scalar-gap-check}
\end{equation}
The gap bound holds for $T\rho_\star\ge C$ and sufficiently small
$\alpha$ and $\mu$.

At time $t$, independently draw
$B_t\sim\operatorname{Bernoulli}(\theta)$. Conditional on
$B_t=1$ and the past, draw $\zeta_t\in\{-1,1\}$ with
\begin{equation}
\Pr(\zeta_t=1\mid\F_{t-1},B_t=1)
=\frac{1+\psi\left(\frac{Lz_t}{H}\right)}2,
\qquad
g_t=B_tA \zeta_t.
\label{eq:signed-bernoulli-oracle}
\end{equation}
Then
\begin{equation}
\E[g_t\mid\F_{t-1}]
=\theta A\psi\left(\frac{Lz_t}{H}\right)
=f'(z_t).
\label{eq:scalar-oracle-unbiased}
\end{equation}
The denominator process depends only on the variables $B_t$, because
$g_t^2=B_tA^2$. Since $|\psi|\le1/2$, both signs have conditional
probability at least $1/4$ whenever $B_t=1$.

\subsection{Displacement from one nonzero observation}

Starting from $m=v=0$, give the algorithm one observation $A$ and
zeros thereafter.  The total absolute displacement of this trajectory is
\begin{equation}
\mathcal D
=\eta\sum_{k=0}^\infty
\frac{\rho_1A\beta_1^k}
{\sqrt{\rho_2A^2\beta_2^k}+\eps}.
\label{eq:single-observation-displacement}
\end{equation}
Set
\begin{equation}
\chi=\frac{L\mathcal D}{H}.
\label{eq:scalar-displacement-ratio}
\end{equation}
Since every denominator in \eqref{eq:single-observation-displacement} is at
least $\eps$,
\begin{equation}
\begin{aligned}
\mathcal D&\le\frac{\eta\rho_1A}{\eps}
\sum_{k=0}^\infty\beta_1^k=\frac{\eta A}{\eps},\\
\chi&=\frac{L\mathcal D}{H}
\le\frac{L\eta A}{H\eps}
=\frac{L\eta}{\theta\eps}=\lambda\frac{\rho_1}{\theta}.
\end{aligned}
\label{eq:single-displacement-upper}
\end{equation}

\subsection{Small-displacement case}

Suppose $\chi\le\chi_0$.

Let $\tau_1,\ldots,\tau_N$ be the times at which $B_t=1$.  Write the first
moment as the sum of the signed contributions from these observations.
At every time $t$,
\begin{equation}
\frac{|m_t|}{\sqrt{v_t}+\eps}
\le\sum_{j=1}^N
\frac{|m_t^{(j)}|}{\sqrt{v_t^{(j)}}+\eps}.
\label{eq:impulse-decomposition}
\end{equation}
Here, for $t\ge\tau_j$,
$m_t^{(j)}=\rho_1A\zeta_{\tau_j}\beta_1^{t-\tau_j}$ and
$v_t^{(j)}=\rho_2A^2\beta_2^{t-\tau_j}$; both contributions are
zero before $\tau_j$. Thus $m_t=\sum_jm_t^{(j)}$ and
$v_t=\sum_jv_t^{(j)}$.
The displayed inequality follows because $|m_t|\le\sum_j|m_t^{(j)}|$
and the full second moment is at least
each nonnegative contribution $v_t^{(j)}$.  Summing over time and extending
each individual displacement contribution to an infinite horizon gives the pathwise bound
\begin{equation}
\sum_{t=1}^T|z_{t+1}-z_t|\le N\mathcal D.
\label{eq:scalar-total-variation}
\end{equation}
Since $\E N=T\theta$, Markov's inequality gives
\begin{align}
\Pr(N\mathcal D\ge W)
&\le\frac{T\theta\mathcal D}{W}
=\chi\frac{T\theta H^2}{\Delta L}\nonumber\\
&=\chi\frac{T\alpha\rho_\star\mu^2\alpha^{2(p-1)/p}R_\star^2}
{T\rho_\star R_\star^2}
=\chi\mu^2\alpha^{1+\frac{2(p-1)}p}.
\label{eq:small-displacement-prob}
\end{align}
Choose $\alpha,\mu$ and then
$\chi_0$ so that \eqref{eq:small-displacement-prob} is at most $1/4$.
On the complementary event, \eqref{eq:scalar-total-variation} and
\eqref{eq:scalar-start} imply
$z_t\ge H/(2L)$ for every $t$, so
$|f'(z_t)|=H/2$. Hence
\begin{equation}
\frac1T\sum_{t=1}^T\E|f'(z_t)|
\ge\frac38H
\ge cR_\star.
\label{eq:small-displacement-lower}
\end{equation}

\subsection{Large-displacement case}

Suppose $\chi>\chi_0$.

Put $q=\rho_2/\theta$. Equation
\eqref{eq:single-displacement-upper} and
\eqref{eq:one-d-parameter-set} imply
\begin{equation}
q=r\frac{\rho_1}{\theta}
\ge\frac{(\inf\mathcal K)\chi_0}{\bar\lambda}
=:q_0>0.
\label{eq:q-lower}
\end{equation}
We first show that a block of length $O(1/\theta)$ captures a fixed
fraction of \eqref{eq:single-observation-displacement}.

\begin{lemma}[Finite-horizon displacement bound]
\label{lem:finite-displacement}
Fix $q_0>0$.  There are constants $B,c>0$, depending only on
$p,\mathcal K,q_0$, with the following property.  Let
$\theta>0$, $\rho_2/\theta\ge q_0$, and
$A=\mu\Sigma\theta^{-\frac{1}{p}}$. Consider a block of length
$\ell=\lceil B/\theta\rceil$.  Suppose its initial second-moment state
satisfies $v\le4\theta A^2$, there is no nonzero observation in the first
quarter, exactly one observation $\zeta A$ in the second quarter, and no other
nonzero observation.  For every time in the last quarter, the positions
obtained with $\zeta=1$ and $\zeta=-1$ differ by at least
$2c\mathcal D(\theta,A)$, where $\mathcal D(\theta,A)$ is defined as in
\eqref{eq:single-observation-displacement}.
The bound is independent of the initial position and first-moment state.
\end{lemma}

\begin{proof}
Let
\[
\mathcal D_n(\theta,A)=\eta\sum_{j=0}^{n-1}
\frac{\rho_1A\beta_1^j}
{\sqrt{\rho_2A^2\beta_2^j}+\eps},
\qquad
\mathcal D(\theta,A)=\lim_{n\to\infty}\mathcal D_n(\theta,A).
\]
Writing $q=\rho_2/\theta$ and using
$\eps=\Sigma\rho_2^{-\frac{2-p}{2p}}$ gives
\begin{equation}
\frac{\sqrt{\rho_2}A}{\eps}
=\mu\rho_2^{\frac{1}{p}}\theta^{-1/p}
=\mu\left(\frac{\rho_2}{\theta}\right)^{1/p}
=\mu q^{1/p}.
\label{eq:displacement-scale-ratio}
\end{equation}
Let
\[
j_0=\left\lceil\frac{2}{\rho_2}
\log_+\left(2\mu q^{\frac{1}{p}}\right)\right\rceil.
\]
Here $\log_+u=\max\{0,\log u\}$. Since
$\beta_2^{j/2}\le e^{-\rho_2j/2}$, for $j\ge j_0$ we have
\[
\frac{\sqrt{\rho_2}A\beta_2^{j/2}}{\eps}
\le\mu q^{1/p}e^{-\rho_2j_0/2}\le\frac12.
\]
For $j\ge j_0$, the square-root term in the denominator is at most
$\eps/2$. Summing only the terms with $j\ge j_0$ gives
\begin{equation}
\mathcal D(\theta,A)
\ge\frac{2\eta\rho_1A}{3\eps}\sum_{j=j_0}^\infty\beta_1^j
=\frac{2\eta A}{3\eps}\beta_1^{j_0}
\ge c\frac{\eta A}{\eps}\beta_1^{j_0}.
\label{eq:displacement-lower-tail}
\end{equation}
Also,
\begin{equation}
\mathcal D(\theta,A)-\mathcal D_n(\theta,A)
\le\frac{\eta\rho_1A}{\eps}\sum_{j=n}^\infty\beta_1^j
=\frac{\eta A}{\eps}\beta_1^n.
\label{eq:displacement-upper-tail}
\end{equation}
Hence $\mathcal D_n(\theta,A)\ge\mathcal D(\theta,A)/2$ whenever
$n\ge j_0+C/\rho_1$. Indeed,
\[
\frac{\mathcal D-\mathcal D_n}{\mathcal D}
\le\frac32\beta_1^{n-j_0}
\le\frac32e^{-\rho_1(n-j_0)}\le\frac12
\]
as soon as $n-j_0\ge(\log3)/\rho_1$.

Both terms in this lower bound for $n$ are $O(1/\theta)$:
\[
j_0\le1+\frac{C(1+\log_+q)}{q\theta}
\le\frac{C}{\theta},
\qquad
\frac1{\rho_1}=\frac{r}{q\theta}
\le\frac{\sup\mathcal K}{q_0\theta}.
\]
The function $(1+\log_+q)/q$ is bounded on $[q_0,\infty)$,
and $\theta=\rho_2/q\le1/(2q_0)$ absorbs the constant $1$ in the bound
for $j_0$. Partition each block into four consecutive sets with endpoints
$\lfloor j\ell/4\rfloor$, $j=0,\ldots,4$.
Choose $B$ large enough that a nonzero observation in the second quarter
precedes every time in the last quarter by at least this many iterations.
Such a choice is possible since $t-s\ge\ell/4-1$ for these two quarters.

Now account for the inherited second moment.  If the observation
occurs at time $s$, then for $j\ge0$ the two sign trajectories have the
same denominator and
\begin{align*}
m_{s+j}^{(\zeta)}&=\beta_1^{j+1}m_{s-1}+\rho_1A\beta_1^j\zeta,\\
v_{s+j}&=\beta_2^j(\beta_2v_{s-1}+\rho_2A^2).
\end{align*}
Since $v_{s-1}\le4\theta A^2$ and $\rho_2/\theta\ge q_0$,
\[
\beta_2v_{s-1}+\rho_2A^2
\le\left(1+\frac{4}{q_0}\right)\rho_2A^2.
\]
The inherited numerator is identical for the two signs and cancels when
their positions are subtracted. The inherited second moment satisfies
\[
\sqrt{v_{s+j}}+\eps
\le\sqrt{1+4/q_0}\,
\left(\sqrt{\rho_2A^2\beta_2^j}+\eps\right).
\]
Since the trajectories agree at time $s$, their separation is therefore
\begin{align*}
|z_t^{(+)}-z_t^{(-)}|
&=2\eta\rho_1A\sum_{j=0}^{t-s-1}
\frac{\beta_1^j}{\sqrt{v_{s+j}}+\eps}\\
&\ge\frac{2\eta\rho_1A}{\sqrt{1+4/q_0}}
\sum_{j=0}^{t-s-1}
\frac{\beta_1^j}{\sqrt{\rho_2A^2\beta_2^j}+\eps}\\
&=\frac{2}{\sqrt{1+4/q_0}}\mathcal D_{t-s}(\theta,A)
\ge\frac{1}{\sqrt{1+4/q_0}}\mathcal D(\theta,A).
\end{align*}
The truncation estimate applies because $t-s$ exceeds the number of
iterations required to capture half of $\mathcal D$. This proves the lemma.
\end{proof}

Partition the horizon into complete blocks of length
$\ell=\lceil B/\theta\rceil$. Since
\[
\E v_t=\beta_2\E v_{t-1}+\rho_2\theta A^2,
\qquad
\E v_t=\rho_2\theta A^2\sum_{j=0}^{t-1}\beta_2^j
=\theta A^2(1-\beta_2^t)\le\theta A^2,
\]
Markov's inequality shows that a block starts with
$v\le4\theta A^2$ with probability at least
$3/4$: for a block starting at step $t_0$,
$\Pr(v_{t_0-1}>4\theta A^2)\le\E v_{t_0-1}/(4\theta A^2)\le1/4$.
The future Bernoulli variables are independent of this event.
Let $I_2$ denote the second quarter of the block.
Increase $B$ so that $\ell\ge8$.
Then $|I_2|\ge\ell/8$, $B\le\ell\theta\le B+1$, and $\theta\le1/2$.
The probability of exactly one nonzero observation in the second quarter
and none elsewhere in the block is bounded by
\begin{equation}
\begin{aligned}
|I_2|\theta(1-\theta)^{\ell-1}
&\ge\frac{\ell\theta}{8}e^{-2\theta(\ell-1)}\\
&\ge\frac B8e^{-2(B+1)}\ge c>0.
\end{aligned}
\label{eq:block-pattern-probability}
\end{equation}
The first inequality uses $\log(1-\theta)\ge-2\theta$.
Thus the joint event consisting of a small initial second moment and
the prescribed observation pattern has probability at least $c>0$.

Fix the history before a block with $v\le4\theta A^2$, and condition on
its Bernoulli pattern and on the time $s$ of the unique observation.
Before $s$ the two sign trajectories coincide, and after $s$ all oracle
observations in this block are zero. The remaining randomness relevant to their
positions is the sign $\zeta$ at time $s$.
For every $t$ in the last quarter, Lemma~\ref{lem:finite-displacement} gives
\begin{equation}
\frac{Lz_t^{(\zeta)}}{H}=C_t-\zeta a_{s,t},
\qquad
a_{s,t}\ge c\frac{L\mathcal D}{H}
\ge c\chi_0,
\label{eq:two-sign-separation}
\end{equation}
where $C_t$ is the midpoint of the two scaled positions and is fixed
under this conditioning. For every $C\in\R$ and $a\ge0$,
\begin{equation}
|\psi(C-a)|+|\psi(C+a)|\ge\min\left\{2a,\frac{1}{2}\right\}.
\label{eq:clip-two-points}
\end{equation}
To verify this inequality, if either $|C-a|$ or $|C+a|$ is at least
$1/2$, its clipped magnitude is $1/2$. Otherwise both terms are unclipped,
and $|C-a|+|C+a|\ge|(C+a)-(C-a)|=2a$.
At time $s$, both signs have conditional probability at least $1/4$.
The future Bernoulli variables are independent of this sign, so
conditioning on the rest of the pattern preserves that probability bound.
Writing $\E_\zeta$ for expectation over these two signs, we therefore have
\begin{align*}
\E_\zeta|f'(z_t)|
&\ge\frac H4\bigl(|\psi(C_t-a_{s,t})|+|\psi(C_t+a_{s,t})|\bigr)\\
&\ge\frac H4\min\{2c\chi_0,1/2\}.
\end{align*}
Integrating over the histories and patterns, whose joint probability is
at least $c>0$, yields $\E|f'(z_t)|\ge cH$ at each time in the
last quarter of a complete block.

There are $M=\lfloor T/\ell\rfloor$ complete blocks, each with at least
$\ell/8$ such times. The assumption $T\theta\ge C$ ensures $T\ge2\ell$
after increasing $C$, since $\ell\le(B+1)/\theta$. Hence
$M\ge T/(2\ell)$, and
\begin{equation}
\frac1T\sum_{t=1}^T\E|f'(z_t)|
\ge\frac{M\ell}{8T}cH\ge\frac{cH}{16}
\ge cR_\star.
\label{eq:large-displacement-lower}
\end{equation}

\subsection{Sample functions and the moment bound}

To construct sample functions for \eqref{eq:signed-bernoulli-oracle},
draw $U_t\sim\operatorname{Unif}[-1,1]$ independently and define, for every
query point $z$,
\begin{equation}
\zeta_t(z)=\operatorname{sign}(\psi\left(\frac{Lz}{H}\right)-U_t),
\qquad
g_t(z)=B_tA \zeta_t(z).
\label{eq:scalar-sample-gradient}
\end{equation}
The function $z\mapsto g_t(z)$ is nondecreasing.  Hence
$F_t(z)=\int_0^zg_t(y)\,dy$ is convex and has $g_t$ as a subgradient
everywhere (and as its derivative except at finitely many points).
Also, $\Pr(U_t<\psi(Lz/H))=(1+\psi(Lz/H))/2$, and therefore
$\E_U \zeta_t(z)=\psi(Lz/H)$.
Independence of $B_t$ and $U_t$ proves \eqref{eq:scalar-oracle-unbiased}.
Finally, $\E[|g_t|^p\mid\F_{t-1}]=\theta A^p$ and
$|f'(z_t)|\le H/2\le H$. Thus
\begin{align}
\E[|g_t-f'(z_t)|^p\mid\F_{t-1}]
&\le2^{p-1}\left(\theta A^p+H^p\right)\\
&\le2^p\theta A^p
=2^p\mu^p\Sigma^p
\le\Sigma^p
\label{eq:scalar-oracle-moment}
\end{align}
after choosing $\mu\le1/2$.
Equations \eqref{eq:small-displacement-lower} and
\eqref{eq:large-displacement-lower} cover all
admissible $(\rho_1,\lambda)$ and prove
Proposition~\ref{prop:one-d}.

\end{document}